\documentclass{article}

\usepackage{arxiv}

\usepackage{natbib}
\usepackage{amsmath}
\usepackage{siunitx}
\usepackage{amssymb}
\usepackage[utf8]{inputenc} % allow utf-8 input
\usepackage[T1]{fontenc}    % use 8-bit T1 fonts
\usepackage{hyperref}       % hyperlinks
\usepackage{url}            % simple URL typesetting
\usepackage{booktabs}       % professional-quality tables
\usepackage{amsfonts}       % blackboard math symbols
\usepackage{nicefrac}       % compact symbols for 1/2, etc.
\usepackage{microtype}      % microtypography
\usepackage{xcolor}         % colors
\usepackage{graphicx}       % graphics
\DeclareSIUnit{\million}{M}

\everypar{\looseness=-1}

\usepackage{etoolbox}
\makeatletter
\patchcmd{\hyper@makecurrent}{%
    \ifx\Hy@param\Hy@chapterstring
        \let\Hy@param\Hy@chapapp
    \fi
}{%
    \iftoggle{inappendix}{%true-branch
        \@checkappendixparam{chapter}%
        \@checkappendixparam{section}%
        \@checkappendixparam{subsection}%
        \@checkappendixparam{subsubsection}%
        \@checkappendixparam{paragraph}%
        \@checkappendixparam{subparagraph}%
    }{}%
}{}{\errmessage{failed to patch}}

\newcommand*{\@checkappendixparam}[1]{%
    \def\@checkappendixparamtmp{#1}%
    \ifx\Hy@param\@checkappendixparamtmp
        \let\Hy@param\Hy@appendixstring
    \fi
}
\makeatletter

\newtoggle{inappendix}
\togglefalse{inappendix}

\apptocmd{\appendix}{\toggletrue{inappendix}}{}{\errmessage{failed to patch}}
\title{Matching the Optimal Denoiser in Point Cloud Diffusion with (Improved) Rotational Alignment}
\newcommand{\R}{\mathbf{R}}
\newcommand{\x}{\mathbf{x}}
\newcommand{\N}{\mathcal{N}}
\renewcommand{\O}{\mathcal{O}}
\newcommand{\EE}{\mathbb{E}}

\newcommand{\Daug}{D_\text{aug}}

\newcommand{\Dperf}{D_\text{perf}}
\newcommand{\Dest}{D_{\text{est}}}
\newcommand{\Dopt}{D^*}
\newcommand{\Raug}{\R_\text{aug}}
\newcommand{\Ropt}{\R^*}
\newcommand{\xz}{x_0}
\newcommand{\pxcy}{{p(x \ | \ y, \sigma)}}
\newcommand{\pxRcy}{{p(x, \R \ | \ y, \sigma)}}
\newcommand{\pRcyx}{{p(\R \ | \ y, x, \sigma)}}
\newcommand{\pRcyxz}{{p(\R \ | \ y, x_0, \sigma)}}
\newcommand{\pRcyRaugx}{{p(\R \ | \ y, \Raug \circ x; \ \sigma)}}
\newcommand{\pycRx}{p(y \ | \ \R, x, \sigma)}

\newcommand{\RR}{\mathbb{R}}
\newcommand{\II}{\mathbb{I}}

\newcommand{\unifR}{u_\R}
\newcommand{\ldenoise}{l_{\text{denoise}}}
\newcommand{\laug}{l_{\text{aug}}}
\newcommand{\lmatch}{l_{\text{match}}}

\newcommand{\lest}{l_{\text{est}}}
\newcommand{\lalignaug}{l_{\text{align-aug}}}

\newcommand{\lnoaug}{l_{\text{no-aug}}}
\newcommand{\norm}[1]{\left\lVert#1\right\rVert}
\newcommand{\btheta}{\boldsymbol\theta}
\DeclareMathOperator*{\argmin}{\arg\!\min}
\DeclareMathOperator*{\argmax}{\arg\!\max}
\DeclareMathOperator{\Tr}{Tr}
\DeclareMathOperator{\SVD}{SVD}
\DeclareMathOperator{\MF}{MF}
\DeclareMathOperator{\diag}{diag}
\DeclareMathOperator{\aug}{Aug}

\title{Matching the Optimal Denoiser in Point Cloud Diffusion with (Improved) Rotational Alignment}

\hypersetup{
pdftitle={Matching the Optimal Denoiser in Point Cloud Diffusion with (Improved) Rotational Alignment},
pdfsubject={cs.LG},
pdfauthor={Ameya Daigavane, YuQing Xie, Bodhi P. Vani, Saeed Saremi, Joseph Kleinhenz, Tess Smidt},
pdfkeywords={kabsch,alignment,diffusion,rotational,SO3,expansion},
}

\author{%
    Ameya Daigavane\thanks{Equal contribution.} \href{https://orcid.org/0000-0002-5116-3075}{\includegraphics[scale=0.06]{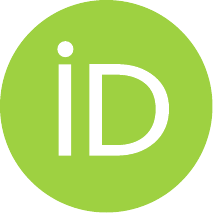}} \\
    Massachusetts Institute of Technology \\
    \texttt{ameyad@mit.edu} \\
    \And
    YuQing Xie\footnotemark[1] \href{https://orcid.org/0000-0002-9949-3700}{\includegraphics[scale=0.06]{orcid.pdf}} \\
    Massachusetts Institute of Technology \\
    \texttt{xyuqing@mit.edu} \\
    \And
    Bodhi P. Vani \href{https://orcid.org/0000-0002-7747-279X}{\includegraphics[scale=0.06]{orcid.pdf}} \\
    Prescient Design, Genentech \\
    \texttt{vanib@gene.com} \\
    \And
    Saeed Saremi \\
    Prescient Design, Genentech \\
    \texttt{saremis@gene.com} \\
    \And
    Joseph Kleinhenz \href{https://orcid.org/0000-0003-3670-0431}{\includegraphics[scale=0.06]{orcid.pdf}} \\ % Placeholder for ORCID
    Prescient Design, Genentech \\
    \texttt{kleinhej@gene.com} \\
    \And
    Tess Smidt \href{https://orcid.org/0000-0001-5581-5344}{\includegraphics[scale=0.06]{orcid.pdf}}\\
    Massachusetts Institute of Technology\\
    \texttt{tsmidt@mit.edu}
}

\begin{document}

\maketitle

\begin{abstract}
Diffusion models are a popular class of generative models trained to reverse a noising process starting from a target data distribution. Training a diffusion model consists of learning how to denoise noisy samples at different noise levels. When training diffusion models for point clouds such as molecules and proteins, there is often no canonical orientation that can be assigned. To capture this symmetry, the true data samples are often augmented by transforming them with random rotations sampled uniformly over $SO(3)$. Then, the denoised predictions are often rotationally aligned via the Kabsch-Umeyama algorithm to the ground truth samples before computing the loss. However, the effect of this alignment step has not been well studied. Here, we show that the optimal denoiser can be expressed in terms of a matrix Fisher distribution over $SO(3)$. Alignment corresponds to sampling the mode of this distribution, and turns out to be the zeroth order approximation for small noise levels, explaining its effectiveness. We build on this perspective to derive better approximators to the optimal denoiser in the limit of small noise. Our experiments highlight that alignment is often a `good enough' approximation for the noise levels that matter most for training diffusion models.
\end{abstract}

\section{Introduction}

Diffusion-based generative models have emerged as a powerful class of generative models for complex distributions in high-dimensional spaces, such as natural images and videos. 

Diffusion models operate on the principle of reversing a noising process by \emph{learning how to denoise}.
Let $p_x$ be our data distribution defined over $\RR^d$. Let $p_y(\cdot ; \sigma)$ be the distribution of $y = x + \sigma \eta$ where $x \sim p_x, \eta \sim \N(0, \II_d)$. $y$  represents a noisy sample at a particular noise level $\sigma$. When the noise level is zero, we have that $p_y(\cdot; \sigma = 0) = p_x$, as expected. On the other end, as $\sigma \to \infty$, the data distribution $p_x$ is effectively wiped out by the noise and $p_y(\cdot; \sigma) \to  \N(0, \sigma^2 \II_d)$. 
As explained by \citet{edm}, the
idea of diffusion models is to randomly sample an initial noisy sample $y_M \sim \N(0, \sigma_M^2 \II_d)$, where $\sigma_M$ is some large enough noise level, and sequentially
denoise it into samples $y_i$ with noise levels $\sigma_M > \sigma_{M - 1} > \cdots > \sigma_0 = 0$ so that at each noise
level $y_i \sim p(y
; \sigma_i)$. Assuming the denoising process has no error, the endpoint $y_0$ of this process will be distributed according to $p_x$.

\begin{figure}[t]
    \centering
    \includegraphics[width=0.8\linewidth]{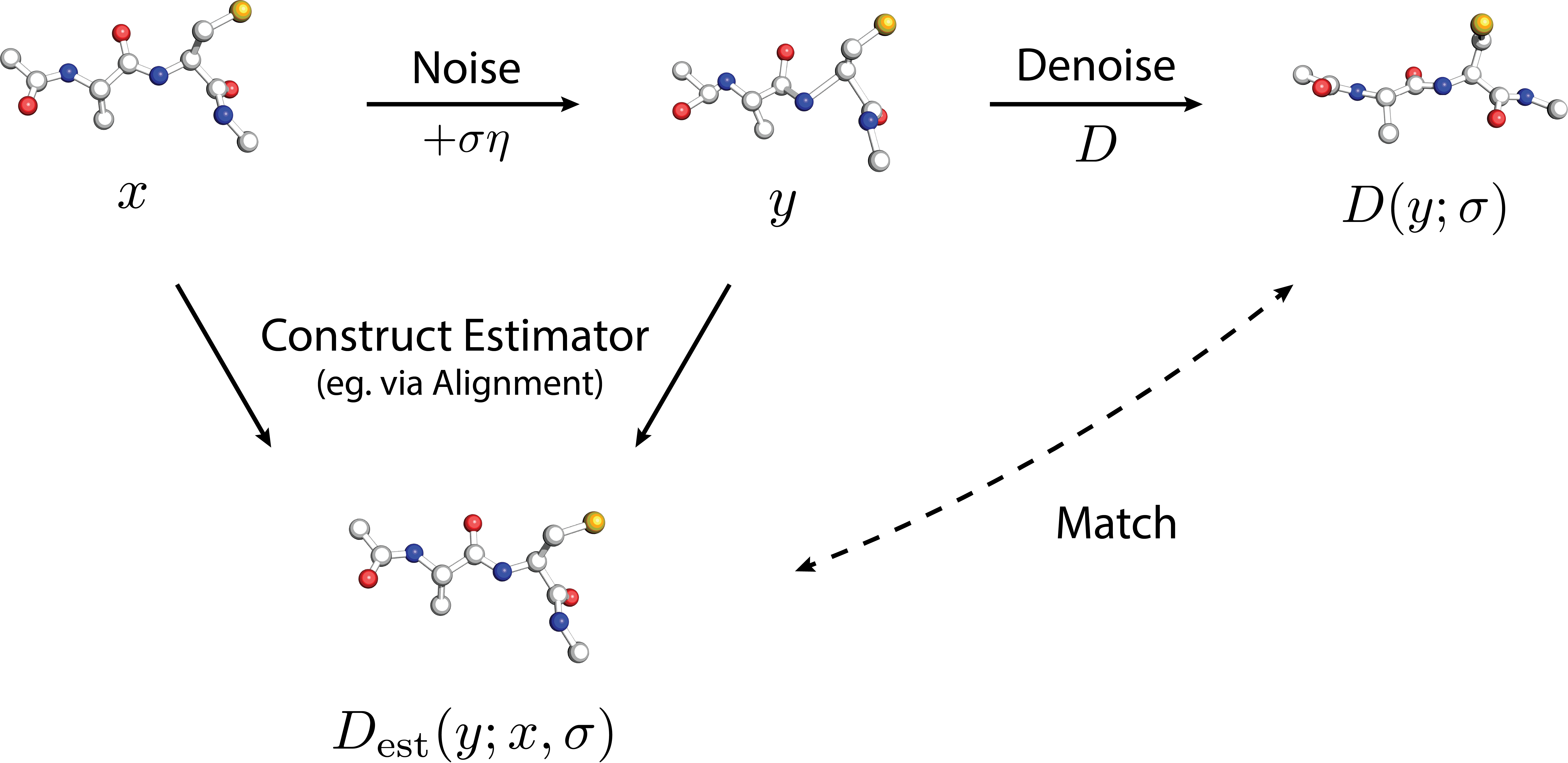}
    \caption{Overview of the training process of a denoising diffusion model, represented by $D$. A sample point cloud $x$ is first noised to give $y$. $D$ denoises $y$ to give a new point cloud $D(y; \sigma)$, which gets matched to an estimator $\Dest(y; x, \sigma)$ of the optimal denoiser $\Dopt$. The usual estimator is $\Dest(y; x, \sigma)  = x$. Here, we show that rotational alignment gives rise to better estimators of $\Dopt$.}
    \label{fig:overview}
\end{figure}

Many schemes \citep{ddim,ddsm,edm} have been built for sampling diffusion models; the details of which are not relevant here. The essential component across these is that one trains a denoiser model $D$ that minimizes the denoising loss $\ldenoise$ across noisy samples at a range of noise levels:
\begin{align}
    \label{eqn:ldenoise}
    \min_D \ldenoise(D) = \min_D  \EE_{x \sim p_x}\EE_{\eta \sim p_\eta}\EE_{\sigma \sim p_\sigma}[\norm{D(x + \sigma \eta; \sigma) - x}^2]
\end{align}
Once learned, the denoiser model $D$ is then used to iteratively sample the next $y_i$ from $y_{i + 1}$ using a numerical integration scheme, such as the DDIM update rule \citep{ddim} (for example):
\begin{align}
    \label{eqn:ddim}
    y_{i} = y_{i + 1} + \left(1 - \frac{\sigma_i}{\sigma_{i + 1}}\right)(D(y_{i + 1}, \sigma_{i + 1}) - y_{i + 1})
\end{align}
We are particularly interested in the setting where the data samples naturally live in three-dimensional space, such as molecular conformations, protein structures and point clouds. 
Often in these settings, there is no canonical 3D orientation that can be assigned to the data samples. Thus, we would like to sample all orientations of the data samples with equal probability. Formally, this means that our `true' data distribution is $\aug[p_x]$, where each sample $x \sim p_x$ has been augmented with uniformly sampled rotations $\R \in SO(3)$.

There are two main approaches to obtain this goal of making the sampled distribution $SO(3)$-invariant (at least approximately):  1) \emph{learning an $SO(3)$-equivariant denoiser $D$} or 2)  \emph{learning with rotational augmentation}. 

Furthermore, to enforce this rotational symmetry, it is common \citep{geodiff,alphafold3,boltz1,jamun,klein2023equivariant,flowmol3} to perform an alignment step (either between $y$ and $x$, or between $D(y; \sigma)$ and $x$) with the Kabsch-Umeyama algorithms \citep{kabsch,umeyama} before computing the loss of denoising.
However, not much is known about the effect of this alignment step. In particular, does alignment introduce bias in the learning objective? If so, can we improve the alignment operation to reduce this bias?
Our paper answers these key questions, and is organized as follows:
\begin{itemize}
    % \item In \autoref{sec:no-perfect-denoiser}, we show that ambiguity over rotations prevents perfect denoising. This motivates the construction of the optimal denoiser for point cloud diffusion. 
    \item In \autoref{sec:optimal-denoiser}, we derive the optimal denoiser for point cloud diffusion. Importantly this denoiser separates into an expectation of the optimal augmented single sample denoiser.
    \item In \autoref{sec:matching}, we show that training a diffusion model is equivalent to matching the single sample optimal denoiser, motivating the construction of better estimators of the optimal denoiser.
    \item In \autoref{sec:alignment}, we show the optimal single sample denoiser involves an expectation of a matrix Fisher distribution over $SO(3)$ and rotation alignment corresponds to a zeroth order approximation using the mode of this distribution. 
    \item In \autoref{sec:asymptotic-expansion}, we build on this insight to obtain better estimators  of the optimal denoiser by approximating the expectation via Laplace's method. These estimators enjoy reduced bias relative to the standard alignment based estimator, at \emph{no additional computational cost}, with numerical evidence in \autoref{sec:numerical}.
    \item In \autoref{sec:practice}, we experiment with these improved estimators to train better diffusion models.
    % In practice, we find that the alignment is a good enough approximation for the expectation over $SO(3)$. 
    At lower noise levels, the effective improvement from using these higher-order correction terms is minimal, suggesting that alignment is already a good enough approximation.
\end{itemize}

% Before we can explain our approach, we need to establish some notation.

\section{Problem Setup}

\textbf{Point Clouds}: Let $x \in \RR^{N \times 3}$ represent a point cloud with $N$ points living in 3D space. The entries in $x$ denote the Cartesian coordinates of each point in the point cloud; each row $x_i^\top$ is the vector of 3D coordinates of the $i$th point. We define $\norm{x}^2 \equiv \sum_{i = 1}^N \norm{x_i}^2$.

Often, the point clouds are associated with some $SO(3)$-invariant features (eg. atomic numbers or charges for atoms in a molecule). These features may be sampled a priori and provided to the denoiser, or undergo their own denoising process with a separate denoiser model \citep{edm-molecules,yim2023fast,yim2024improved,multiflow}. Our analysis is not affected in either case, so we omit these features from further discussion.

\textbf{Rotations}: $SO(3)$ refers to the group of rotations in three dimensions. Since we are working with point clouds which are sets of vectors, it is natural to consider the representation of rotations as rotation matrices $\R \in \RR^{3 \times 3}$. Thus, the action of a rotation $\R$ on the point cloud $x = [x_i^\top]_{i = 1}^N$ is $\R \circ x$ obtained by rotating the coordinates of each point by $\R$ independently:
\begin{align}
    \R \circ x \equiv x\R^\top \equiv [x_i^\top \R^\top]_{i = 1}^N.
\end{align}
Note that rotation matrices are orthonormal: $\R^\top\R = \II_3$. Further, the group action is associative:
\begin{align}
    \R_1\R_2 \circ x =
    \R_1 \circ (\R_2 \circ x).
\end{align}
\textbf{$SO(3)$-Invariance}: Let $p$ be a distribution over $\RR^{N \times 3}$. We say that $p$ is $SO(3)$-invariant if:
\begin{align}
    p(\R \circ x) = p(x) \quad \text{for all} \ \R \in SO(3), x \in \RR^{N \times 3}.
\end{align}
Since there is no way to appropriately normalize a translation-invariant $p_x$, we center point clouds $x$ such that $\sum_{i = 1}^N x_i = \Vec{0}$, as is commonly done in the literature. This operation does not change our analysis.

\textbf{Rotational Augmentation}: 
The simple (yet approximate) approach to obtain an equivariant distribution is to augment the data distribution $p_x$ with randomly sampled rotations. In particular, we sample $x \sim p_x$ from our data distribution $p_x$, $\R \sim \unifR$ from the uniform distribution $\unifR$ over $SO(3)$ as defined by the Haar measure, and return $\R \circ x \sim \aug[p_x]$, which is defined as:
\begin{align}
    \aug[p_x](x')=\int_{SO(3)}p_x(\R^{-1} \circ x')\unifR(\R)\mathrm{d}\R.
\end{align}
By construction, $\aug[p_x](\R \circ x) = p_x(x)\unifR(\R)$ for any $\R$.
A simple proof (\autoref{sec:augmentation-makes-invariant}) shows that $\aug[p_x]$ is always $SO(3)$-invariant.
When training with rotational augmentation,\footnote{We could have also denoted the augmentated input to $D$ as $\R \circ x + \sigma \eta$, which is equivalent in expectation due to the isotropy of the Gaussian distribution.}
the loss becomes:
\begin{align}
    \label{eqn:laug}
    \min_D \laug(D) = \min_D \EE_{\R \sim \unifR}\EE_{x \sim p_x}\EE_{\eta \sim p_\eta}\EE_{\sigma \sim p_\sigma}[\norm{D(\R \circ (x + \sigma \eta); \sigma) - \R \circ x}^2].
\end{align}

% Our analysis here focuses on $SO(3)$-equivariant denoising, defined in the next section. However, the `mean-centered point cloud assumption' does not affect any of the analysis. \todo{Check this, because I think in practice we mean-center $y$ as well. I think it should not matter because our alignment mods out translations as well.} Our analysis also follows nearly identically for improper rotations as well. Thus, our results for $SO(3)$-equivariant denoising also apply to $SE(3)$-equivariant and $E(3)$-equivariant denoising as well. \todo{and check this.}

% \textbf{Denoisers}: A denoiser $D: \RR^{N \times 3} \to \RR^{N \times 3}$ takes in a noisy point cloud and predicts a `denoised' point cloud.

\textbf{$SO(3)$-Equivariant Denoisers}:
An $SO(3)$-equivariant denoiser $D$ commutes with all rotations $\R$. Formally:
\begin{align}
    D(\R \circ (x + \sigma \eta)) = \R \circ D(x + \sigma \eta)
\end{align}
for all point clouds $x$, noise $\eta$, noise levels $\sigma$ and rotations $\R$.

% \textbf{Noise Distribution}:
% We consider the usual multivariate Gaussian distribution $p_\eta = \N(0, \sigma^2 \II_{N \times 3})$ over noise $\eta \in \RR^{N \times 3} $. Here, $\sigma$ represents the standard deviation of each entry in $\eta$. We do use the isotropy of the Gaussian to simplify some integrals, but the techniques here can be generalized to arbitrary noise distributions. Hence, our results can be easily generalized to the general frameworks of flow-matching and stochastic interpolants as well.

In \autoref{sec:equivariant-learns-invariant}, we show that, given a $SO(3)$-invariant initial distribution, the result of diffusion sampling with a $SO(3)$-equivariant denoiser is a $SO(3)$-invariant distribution. In our case, our initial noise distribution is an isotropic multivariate Gaussian, so the conditions are satisfied. Further, for an equivariant denoiser, data augmentation has no effect (\autoref{sec:augmentation-equivariant}): $\laug(D) = \ldenoise(D)$ for an equivariant $D$. 
% This implies that the perfect equivariant denoiser will sample from $\aug[p_x]$ when trained to minimize $\ldenoise$ over samples from $p_x$.

In \autoref{sec:no-perfect-denoiser-rigorous}, we show that there is no perfect denoiser under rotational augmentation, implying that $\laug$ has a non-zero minimum. 
To summarize, the fundamental issue is that there is an ambiguity with respect to which orientation $\R \circ x$ to denoise to.
This inspires the idea of alignment to simply cancel out the effect of any such rotation. However, to analyze the alignment step, we first need to characterize the form of the optimal denoiser $\Dopt$ which minimizes $\laug$. 
% Our analysis  reveals a deeper connection between rotational alignment and the optimal denoiser $\Dopt$. 

% When restricted to equivariant denoisers, this implies that $\ldenoise$ has the same non-zero minimum.

\section{The Optimal Denoiser}
\label{sec:optimal-denoiser}
Having established that there does not exist a perfect denoiser, we can now ask about the optimal denoiser $\Dopt$ obtaining the minimum possible loss $\laug(\Dopt) = \min_D \laug(D) > 0$. Here, we adapt the derivation performed in \citet{edm}.

% \subsection{Quick optimal denoiser derivation}

% We can write general denoising loss as
% \begin{align*}
%     l(D) = & \EE_{x \sim p_x}\EE_{\eta \sim p_\eta}\EE_{\sigma \sim p_\sigma}[\norm{D(x + \sigma \eta; \sigma) - x}^2]\\
%     = & \EE_{x, \eta, \sigma \sim p_{x,\eta,\sigma}}[\norm{D(x + \sigma \eta; \sigma) - x}^2]
% \end{align*}
% Now do a change of variables so instead of $x,\eta,\sigma$ we instead have $x,y,\sigma$ where $y=x+\sigma\eta$. Then this equivalently becomes
% \begin{align*}
%     l(D) = & \EE_{x, y, \sigma \sim p_{x,y,\sigma}}[\norm{D(y; \sigma) - x}^2]\\
%      = & \EE_{y,\sigma\sim p_{y,\sigma}}\EE_{x, \sim p_{x|y,\sigma}}[\norm{D(y; \sigma) - x}^2].
% \end{align*}
% It is now obvious that we can just optimize the value of each $D(y,\sigma)$ individually for $\EE_{x, \sim p_{x|y,\sigma}}[\norm{D(y; \sigma) - x}^2]$ and it is not hard to show that this is minimized for $D^*(y;\sigma)=\EE_{x\sim p_{x|y,\sigma}}[x]$.

\subsection{The Optimal Denoiser in the Single Sample Setting}
\label{sec:single-sample-optimal}
We first state the optimal denoiser $\Dopt$ in the single sample case, where $p_x(x) = \delta(x - \xz)$. 
Let $\Dopt(y; \xz, \sigma)$ be the optimal denoiser for $y$ conditional on a particular $\xz$ at a noise level $\sigma$, which we term the \emph{optimal conditional denoiser}. Then:
\begin{align}
    \label{eqn:optimal-conditional-denoiser}
    \Dopt(y; \xz, \sigma) = \frac{\EE_{\R \sim \unifR} [\N(y; \R \circ \xz, \sigma^2 \II_{N \times 3}) \R \circ \xz ]}{\EE_{\R \sim \unifR} [\N(y; \R \circ \xz, \sigma^2 \II_{N \times 3})]}
    &=
    \EE_{\R \sim \pRcyxz} [\R \circ \xz]
\end{align}
as we prove in \autoref{sec:proof-single-sample-optimal}. Above:
\begin{align}
    \label{eqn:induced-distribution-R}
    \pRcyxz = \frac{\N(y; \R \circ \xz, \sigma^2 \II_{N \times 3}) \unifR(\R)}{\int_{SO(3)} \N(y; \R' \xz, \sigma^2 \II_{N \times 3}) \unifR(\R') d\R'}
\end{align}
Thus, we see that the optimal conditional denoiser essentially corresponds to an expectation over different orientations $\R \circ x$. Importantly, it turns out that the optimal conditional denoiser is $SO(3)$-equivariant:
\begin{align}
    \Dopt(\R \circ y; \xz, \sigma) = \R \circ \Dopt(y; \xz, \sigma)
\end{align}
We provide a short proof using the invariance of the Haar measure in \autoref{sec:optimal-denoiser-equivariant}. Further, the optimal conditional denoiser is invariant under rotations $\Raug$ of the conditioning $x$: 
\begin{align}
    \label{eqn:optimal-conditional-equivariance}
    \Dopt(y; \Raug x, \sigma) = \EE_{\R \sim \pRcyRaugx}[
    \R \circ (\Raug \circ x)] = \Dopt(y; x, \sigma)
\end{align}

\subsection{The Optimal Denoiser in the General Setting}
\label{sec:general}

In the general setting where $p_x$ is arbitrary, the optimal denoiser $\Dopt(y; \sigma)$ is an expectation over $x \sim \pxcy $ of the optimal conditional denoiser $\Dopt(y; x, \sigma)$, as we prove in \autoref{sec:proof-multi-sample-optimal}:
\begin{align}
\Dopt(y; \sigma) &= \EE_{x, \R \sim \pxRcy}[\R \circ x] \nonumber\\
  &= \EE_{x \sim \pxcy}\EE_{\R \sim \pRcyx}[\R \circ x]
  = \EE_{x \sim \pxcy}[\Dopt(y; x, \sigma)]\label{eqn:opt-denoiser}
\end{align}
It follows from the $SO(3)$-equivariance of $\Dopt(y; x, \sigma)$ that $\Dopt(y; \sigma)$ is also $SO(3)$-equivariant.

% Thus, to approximate the optimal denoiser, we can develop approximations of the optimal conditional denoiser.
% Given this form of the optimal denoiser, we now 
Given that there is an analytic expression of the optimal denoiser, a natural question arises. Instead of minimizing the usual denoising loss (\autoref{eqn:ldenoise}), can we instead try to match the optimal denoiser? Indeed, we shall see that these approaches are actually identical.

\section{Matching the Optimal Denoiser}
\label{sec:matching}
Here, we motivate why we would want to match the optimal denoiser $\Dopt$. To do so, we first consider matching to some general estimator $\Dest(y;x,\R,\sigma)$ potentially dependent on all the random variables. 
Given any estimator $\Dest(y;x,\R,\sigma)$ we want to match to $D$, we can define a matching loss by:
\begin{align}
    \label{eqn:l_est}
    \lest(D; \Dest) &=  \EE_{\R \sim \unifR}\EE_{x \sim p_x}\EE_{\eta \sim p_\eta}\EE_{\sigma \sim p_\sigma}[\norm{D(\R \circ (x + \sigma \eta); \sigma) - \Dest(\R \circ (x + \sigma \eta); x,\R,\sigma)}^2]\nonumber\\
    &= \EE_{\R \sim \unifR}\EE_{x \sim p_x}\EE_{\sigma \sim p_\sigma}\EE_{y \sim p(y|x,\sigma,\R)}[\norm{D(y; \sigma) - \Dest(y; x,\R,\sigma)}^2]\nonumber\\
    &= \EE_{\sigma \sim p_\sigma}\EE_{y\sim p(y|\sigma)}\EE_{x \sim \pxcy}\EE_{\R \sim \pRcyx}[\norm{D(y; \sigma) - \Dest(y; x, \R, \sigma)}^2]
\end{align}
after identifying $y \equiv \R \circ (x + \sigma \eta)$.
As we prove in \autoref{sec:averaging-an-estimator}, averaging an estimator $\Dest$ over $\R \sim \pRcyx$ to obtain a new estimator $\EE_{\R \sim \pRcyx}[\Dest]$ gives us an equivalent loss from a minimization perspective: $\lest(D; \EE_{\R \sim \pRcyx}[\Dest]) = \lest(D; \Dest) + C$, where $C$ is a constant that does not depend on $D$. The same reasoning applies to averaging an estimator $\Dest$ over $x \sim \pxcy$ to obtain a new estimator $\EE_{x \sim \pxcy}[\Dest]$.

Setting $\Dest(y;x,\R,\sigma) \equiv \R\circ x$ recovers $\laug$. Now, for this choice of $\Dest$, $\EE_{\R \sim \pRcyx}[\Dest]$ corresponds to the \emph{optimal conditional denoiser}. Further, $\EE_{x \sim \pxcy}\EE_{\R \sim \pRcyx}[\Dest]$ corresponds to the \emph{optimal denoiser}. Thus, from the previous arguments, optimal denoiser matching is \emph{equivalent} to the standard denoising loss.

% By choosing optimal denoisers, it turns out the matching loss is equivalent to $\laug$ up to a constant. Further, optimal denoisers eliminate dependence on some random variables, decreasing the sampling error in estimating the expected values.
% Given an estimator $\Dest(y;x,\R,\sigma)$, averaging 
% \autoref{sec:averaging-reduces-mse} shows that averaging $\Dest$ over $\R \sim \pRcyx$ to obtain a new estimator $\EE_{\R \sim \pRcyx}[\Dest]$
% is equivalent to minimizing \autoref{eqn:ldenoise} in expectation. This motivates the construction of estimators $\Dest$ for $\Dopt$.

The advantage of averaging, however, comes from a practical perspective. In practice, we can only estimate this loss through sampling which introduces a sampling error for each random variable (ie, $\x, \R$) in the argument. By averaging, we remove the dependence of the argument on the random variable being averaged over and reduce this sampling error.

Performing the averaging over $x$ is tricky, because the expectation over $\pxcy$ requires access to $p_x$. Approximating this expectation over a finite set of $x$ can suffer from overfitting \citep{vastola,kadkhodaie2024generalization,li2023on}. \citet{niedoba2024nearestneighbourscoreestimators} tried to address this by building estimators based on the nearest-neighbor approximation. However, the averaging over $\R$ to give the optimal conditional denoiser is indeed feasible, as we discuss next.

%  By setting $\Dest(y;x, \R, \sigma)=D^*(y;\sigma)$, we would actually have $\lest(D)=\laug(D)+C$ where $C$ is some constant not dependent on $D$, which we prove in \autoref{sec:optimal-denoiser-matching}. In principle, we can match to the optimal denoiser instead. However, this is infeasible since the optimal denoiser has an expectation over $\pxcy$ which requires access to $p_x$. 
% Instead, we can try to only remove the dependence on $\R$ by matching to $\Dest(y;x, \R, \sigma)=D^*(y;x,\sigma)$. It turns out in this case we also have $\lest(D)=\laug(D)+C$ where $C$ is some constant. But the single sample optimal denoiser is tractable. Hence we can use it to remove the variance error associated from sampling $\R \sim p(\R|y,x,\sigma)$.

In conclusion, rather than optimizing $\laug$, we can instead match to the optimal conditional denoiser:
\begin{align}
    \lest(D; D^*(y, x, \sigma)) &= \lmatch (D) =\EE_{\sigma \sim p_\sigma}\EE_{y\sim p(y|\sigma)}\EE_{x \sim \pxcy}[\norm{D(y; \sigma) - D^*(y; x, \sigma)}^2]. \label{eqn:lmatch}
\end{align}

\section{The Matrix Fisher Distribution on $SO(3)$}
\label{sec:mf-distribution}

% While normally the optimal denoiser is intractable since we do not have access to $p_x$, in the case of rotation augmentation we see from \autoref{eqn:opt_denoiser} our optimal denoiser splits into two parts. One part depends on $p_x$ but the other corresponds to \autoref{eqn:optimal-conditional-denoiser} the augmented single sample case. Importantly, we can explicitly describe the latter.

Here, we connect the optimal conditional denoiser to the matrix Fisher distribution. Recall our expression of $D^*(y;x,\sigma)$ involves an expectation over $\pRcyx$. Some simple algebraic manipulation (\autoref{sec:matrix-fisher-connection}) shows us that:
\begin{align}
    \label{eqn:pr-mf}
    \pRcyx \propto \exp\left(-\frac{\norm{y - \R \circ x}^2}{2\sigma^2}\right)  \propto \exp\left(\Tr\left[\frac{y^\top x}{2\sigma^2}\R\right]\right)
    % \quad
    % \implies \quad \pRcyx = \MF\left(\R; \frac{y^\top x}{\sigma^2}\right)
\end{align}
In particular, the distribution $\pRcyx$ belongs to a well-studied family of distributions called the matrix Fisher distribution over $SO(3)$ \citep{mohlin2020probabilistic,lee2018bayesian}. This distribution is parametrized by a $3 \times 3$ matrix $F$:
\begin{align}
    \label{eqn:mf-definition}
   \MF(\R; F) = \frac{\exp{\Tr[F^\top\R]}}{Z(F)} 
\end{align}
where $Z(F)$ is the partition function, ensuring that the distribution is normalized over $SO(3)$. 
Thus, conditional on $y, x$ and $\sigma$, $\R$ is distributed according to a matrix Fisher distribution with $F = \frac{y^\top x}{\sigma^2} \in \RR^{3\times 3}$. We can write:
\begin{align}
    D^*(y;x,\sigma)=\EE_{\R\sim \MF\left(\R; \frac{y^\top x}{\sigma^2}\right)}[\R\circ x]=\EE_{\R\sim \MF\left(\R; \frac{y^\top x}{\sigma^2}\right)}[\R]\circ x.
\end{align}
Therefore computing the optimal denoiser involves computing the first moment of a matrix Fisher distribution.

\section{Rotational Alignment}
\label{sec:alignment}

It turns out rotation alignment gives a very good approximation for the first moment. Assuming $x,y$ have no rotational self symmetries (which is generically the case), then $\MF\left(\R; \frac{y^\top x}{\sigma^2}\right)$ is a unimodal distribution. In fact, for small $\sigma$ this distribution becomes very sharply peaked, so the mode becomes a good approximation of the first moment.

We can see the mode of the distribution satisfies:
\begin{align}
    \label{eqn:alignment}
    \R_{\mathrm{mode}} &= \argmax_{\R \in SO(3)} \MF\left(\R; \frac{y^\top x}{\sigma^2}\right) 
    = \argmax_{\R \in SO(3)} \Tr\left[\frac{y^\top x}{2\sigma^2}\R\right]
    =\argmin_{\R\in SO(3)}-\Tr[y^\top x\R]\nonumber\\
    &= \argmin_{\R \in SO(3)} \frac{1}{2}(\Tr[y^\top y]-2\Tr[y^\top x\R]+\Tr[\R^\top x^\top x \R)=\argmin_{\R\in SO(3)}||y-\R\circ x||^2\\
    &= \Ropt(y,x)
\end{align}
where $\Ropt(y, x)$ is the optimal rotation for aligning $x$ to $y$, exactly the rotation matrix returned by the Kabsch and Proscrutes alignment algorithms.

\begin{figure}[h]
\centering\includegraphics[width=0.4\linewidth]{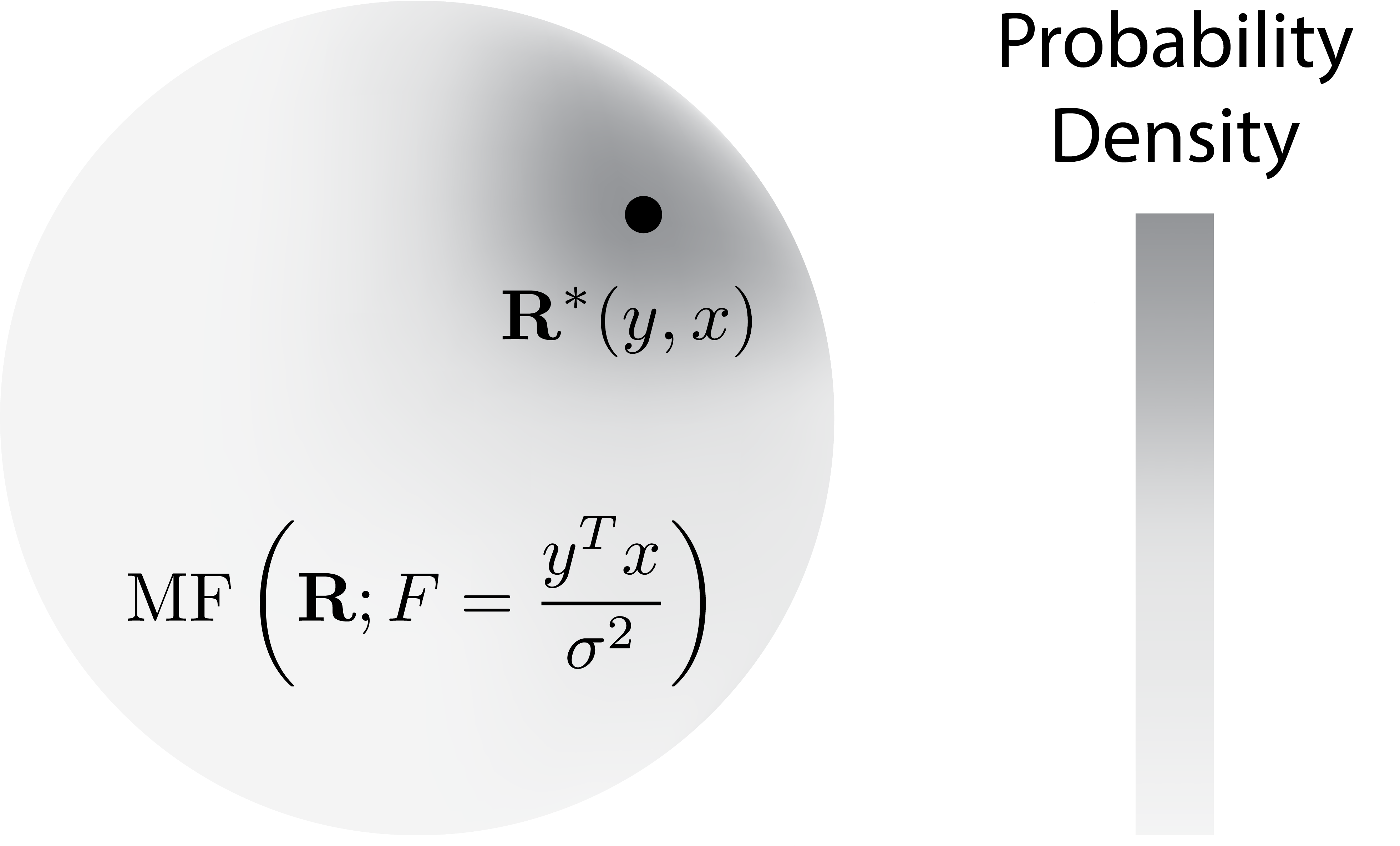}
    \caption{A depiction of the unimodal $\MF(\R; F)$ over $SO(3)$, highlighting the mode $\Ropt(y, x)$. As $\sigma$ decreases, the distribution becomes more peaked around $\Ropt(y, x)$.}
    \label{fig:mf}
\end{figure}

% For simplicity, we assume that the optimization in \autoref{eqn:alignment} has a unique minima, which is usually the case unless $x$ or $y$ have self-symmetries (i.e. if there exist rotations which keep $x$ or $y$ unchanged). 

This marks our first insight: alignment allows us to approximate $\EE_\R[\R]\approx \Ropt(y,x)$ in the optimal single sample denoiser (\autoref{eqn:optimal-conditional-denoiser}) so
\[D^*(y;x,\sigma)=\EE_\R[\R]\circ x \approx \Ropt(y,x)x.\]
% In particular, an estimator for $\Dopt(y; x, \sigma)$ can be constructed by computing the mode $\Ropt(y, x)$ of $\pRcyx$, and returning $\Ropt(y, x) \circ x$. 

Substituting $\Ropt(y,x)x$ for $\Dest$ in \autoref{eqn:l_est}, we obtain
\begin{align}
    \lalignaug(D)=\EE_{\R \sim \unifR}\EE_{x \sim p_x}\EE_{\eta \sim p_\eta}\EE_{\sigma \sim p_\sigma}[\norm{D(\R \circ (x + \sigma \eta); \sigma) - \Ropt(y,x)\circ x}^2]
\end{align}
exactly the loss used for rotation alignment!

Can better estimators for $\Dopt$ be constructed by better approximations of $\EE_\R[\R]$? Yes! In \autoref{sec:asymptotic-expansion}, we show that there exist better approximators of $\EE_\R[\R]$ with no additional asymptotic runtime cost over alignment.

\section{Approximating the Optimal Denoiser via an Asymptotic Expansion of the Matrix-Fisher Distribution}
\label{sec:asymptotic-expansion}

Here, we derive additional correction terms to the moment $\EE_\R[\R]$ in the limit of $\sigma\to0$. These correction terms can be implemented with no additional cost.

% Essentially, we need to compute the expected rotation $\EE_{\R \sim \MF(\R; F)}[\R]$ of a Matrix-Fisher distribution $\MF(\R; F)$. In the limit of small noise level $\sigma$, we end up with a very sharply peaked distribution and rotational alignment gives a very good estimate of the expected rotation. We characterize how good this approximation is via \emph{Laplace's method for integrals}. We first develop the theory for general $F$, and then specialize for $\pRcyx$ with $F = \frac{y^\top x}{\sigma^2}$, as we found in \autoref{eqn:pr-mf}. 

\subsection{For a General Matrix Fisher Distribution}

We provide a high level summary of the method used to derive additional correction terms. More detail can be found in \autoref{sec:laplace_details}.

We note that the partition function $Z(F)=\int_{SO(3)}\exp(\Tr[F^\top\R])\mathrm{d}\R$ gives all the information we need. In particular, taking derivatives of $Z(F)$ allows us to calculate any necessary moments. For example, 
\begin{align}
    \EE_{\R \sim \MF(\R; F)}[\R] = \frac{\int_{SO(3)}\R \exp(\Tr[F^\top\R])\mathrm{d}\R}{Z(F)} = \frac{\frac{d}{dF} Z(F)}{Z(F)} = \frac{d}{dF} \ln Z(F)
\end{align}
using the trace derivative identity: $\frac{d}{dF} \Tr[F^\top \R] = \R$. To control peakedness, we replace $F$ with $\lambda F$ so that as $\lambda\to\infty$, the distribution becomes a delta function centered at the mode.

Next, recall that in Kabsch alignment we use SVD to decompose where $U,V\in SO(3)$ and $S=\diag[s_1,s_2,s_3]$ where $s_1\geq s_2\geq |s_3|$. It turns out that $Z(\lambda F)=Z(\lambda S)$ so we restrict to only expanding $Z$ for diagonal $S$. To approximate $Z(\lambda S)$, we use Laplace's method.

First, we choose to use the exponential map parameterization $\R(\theta_x,\theta_y,\theta_z)=\exp(\theta_x R_x + \theta_y R_y + \theta_z R_z$. Next, we Taylor expand the argument $\Tr[S^\top\R(\btheta)]$ around $\btheta=0$. It is not hard to check that $\II$ maximizes this so there is no first order term. Hence taking up to second order terms, we obtain some $\exp(\lambda A_0(S)+\lambda \btheta^\top A_2(S)\btheta)$ which can be interpreted as a Gaussian because $A_2$ must be negative definite since $\II$ is the maximum. In some sense this captures the peak of the distribution and we can write 
\[\exp(\Tr[\lambda S^\top\R(\btheta])=\exp(\lambda A_0(S)+\lambda \theta^\top A_2(S)\theta)B(\btheta,S,\lambda)\]

Next, we can perform the usual Taylor expansion of $B(\btheta,S,\lambda)\mu(\btheta)$, the remaining terms in the integral where $\mu(\btheta)$ is the corresponding Haar measure. We also do this around $\btheta=0$ because only the neighborhood around the peak matters. Finally, we note that as $\lambda\to\infty$, the width of the peak described by $\exp(\lambda A_0(S)+\lambda \theta^\top A_2(S)\theta)$ decreases. Hence, replacing the domain $\{|\btheta|<\pi\}$ with the larger domain $\{\btheta \in \RR^3\}$ not gives us a good approximation for each of the expanded terms, but also gives us Gaussian integrals which are analytically evaluable.

Finally, we obtain an expression of the form:
\begin{align}Z(\lambda S)=N(S,\lambda)\left(1+L_1(S)\frac{1}{\lambda}+L_2(S)\frac{1}{\lambda^2}+L_3(S)\frac{1}{\lambda^3}+\ldots\right)\end{align}
where $N(S,\lambda)$ is a normalization term. The corresponding expected rotation can be computed as:
\begin{align}
    \EE_{\R\sim\MF(\R;\lambda S)}[\R]  &= \frac{1}{\lambda}\diag\left[\frac{\partial \ln Z(\lambda S)}{\partial s_1},\frac{\partial \ln Z(\lambda S)}{\partial s_1},\frac{\partial \ln Z(\lambda S)}{\partial s_3}\right] \nonumber\\
&= \II+C_1(S)\frac{1}{\lambda}+C_2(S)\frac{1}{\lambda^2}+\ldots
    \label{eqn:expected-R-expand}
\end{align}
For an arbitrary $F'=USV^\top$, we would then have:
\begin{align}
    \EE_{\R\sim\MF(\R;\lambda F')}[\R]=U\EE_{\R\sim\MF(\R; \lambda S)}[\R]V^\top.
\end{align}

\subsection{For the specific $F = \frac{y^\top x}{\sigma^2}$}
\label{sec:approximators}

We are specifically interested in the case where $F = \frac{y^\top x}{\sigma^2}$, as $\pRcyx = \MF(\R;  \frac{y^\top x}{\sigma^2} )$.
From the Kabsch algorithm, we have that $\Ropt(y, x) = UV^\top$ where $U, S, V^\top = \SVD(y^\top x) = \SVD(F')$ with $\det(U) = \det(V) = 1$ and $S = \diag[s_1, s_2, s_3]$ where $s_1 \geq s_2 \geq |s_3|$.

Following the procedure outlined in \autoref{sec:asymptotic-expansion}, we used Mathematica \citep{Mathematica} to find the coefficients in \autoref{eqn:expected-R-expand} explicitly:
% Our claim is that:
% \begin{align}
%     \label{eqn:expansion-expectation}
%     \EE_{\R \sim \pRcyx}[\R] &= U S' V^\top \\
%     S' &= \II_{3\times3} + 
%     \sigma^2 C_1(S) + \sigma^4 C_2(S) + \O\left({\sigma^5}\right)
% \end{align}
% with:
\begin{align}
    C_1(S) &= -\frac{1}{2}\diag\left[\frac{1}{s_1 + s_2} + \frac{1}{s_1 + s_3}, \frac{1}{s_2 + s_1} + \frac{1}{s_2 + s_3}, \frac{1}{s_3 + s_1} + \frac{1}{s_3 + s_2}\right] \\
    C_2(S) &= -\frac{1}{8}\diag\left[\frac{1}{(s_1 + s_2)^2} + \frac{1}{(s_1 + s_3)^2}, \frac{1}{(s_2 + s_1)^2} + \frac{1}{(s_2 + s_3)^2}, \frac{1}{(s_3 + s_1)^2} + \frac{1}{(s_3 + s_2)^2}\right]
\end{align}
These represent the first-order and second-order correction terms respectively. 
This approximation is exact in the limit $\sigma \to 0$.

\autoref{eqn:expected-R-expand} allows us to approximate the optimal conditional denoiser as:
\begin{align}
    \label{eqn:optimal-denoiser-expectation}
    \Dopt(y; x) &= \EE_{\R \sim \pRcyx}[\R \circ x] = 
    \EE_{\R \sim \pRcyx}[\R] \circ x \\ 
    \label{eqn:expansion-denoiser}
    &= (\Ropt(y, x) + \sigma^2 B_1(y, x) + \sigma^4 B_2(y, x)) \circ x + \O(\sigma^5).
\end{align}
where $B_1(y, x) = UC_1(S)V^\top$ and $B_2(y, x) = UC_2(S)V^\top$.

Thus, alignment corresponds to the \emph{zeroth-order} (in $\sigma$) approximation of $\Dopt(y; x, \sigma)$. From \autoref{eqn:expansion-denoiser}, we define the successive \emph{first-order} and \emph{second-order} approximations to the optimal denoiser:
\begin{align}
    \label{eqn:order-approx-0}
    \Dopt_0(y; x, \sigma) &= \Ropt(y, x) \circ x \\
    \label{eqn:order-approx-1}
    \Dopt_1(y; x, \sigma) &= (\Ropt(y, x)  + \sigma^2 B_1(y, x)) \circ x \\
    \label{eqn:order-approx-2}
    \Dopt_2(y; x, \sigma) &= (\Ropt(y, x) + \sigma^2 B_1(y, x) + \sigma^4 B_2(y, x)) \circ x
\end{align}
Our improved estimators $\Dopt_1$ and $\Dopt_2$ have reduced bias relative to the usual alignment-based $\Dopt_0$.
Importantly, these improved estimators can be constructed at no additional computational cost to the standard Kabsch alignment, since $U$ and $V$ have already been computed. They simply correspond to adjusting the rotation $\Ropt$ before multiplying with $x$.

\section{Numerical Error in Approximations of the Optimal Denoiser}
\label{sec:numerical}

\begin{figure}[h]
    \centering
    \includegraphics[width=0.8\textwidth]{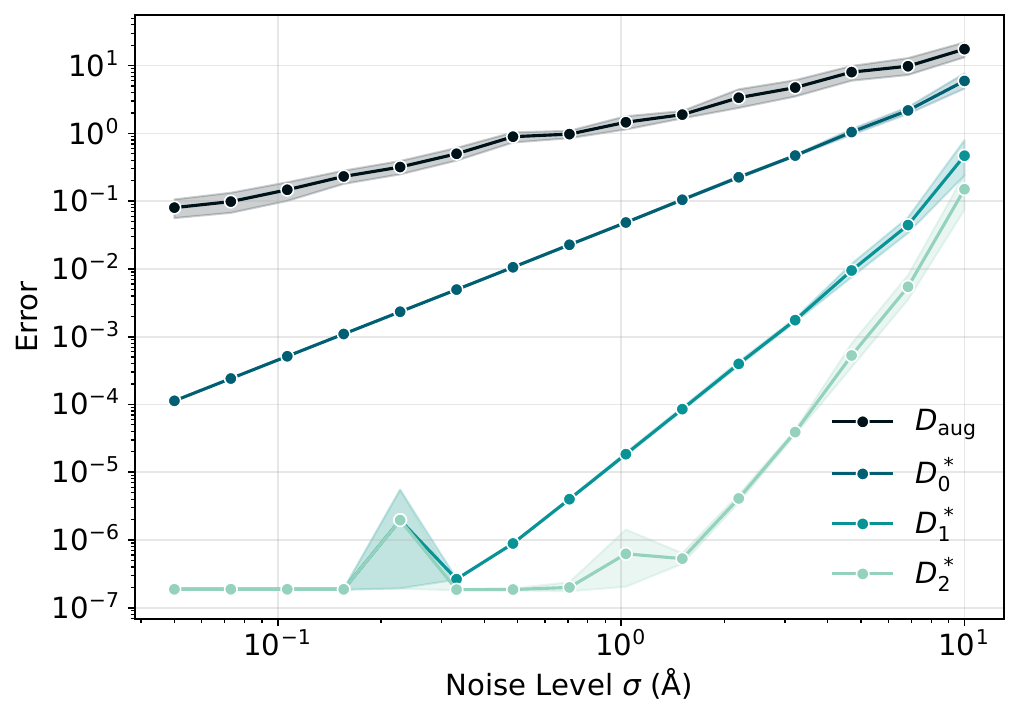}
    \caption{Mean-squared error relative to the optimal denoiser $\Dopt(y; x)$ as a function of $\sigma$. $x$ here is a randomly chosen conformation of the \texttt{AEQN} tetrapeptide from the \textsc{Timewarp 4AA-Large} dataset, and $y$ is sampled as $x + \sigma \eta$. 
    % The optimal denoiser is estimated numerically using Mathematica's integrator over $SO(3)$. 
    % The gray region indicates the values of $\sigma$ for which the resolution of the finite quadrature grid is too small to compute the optimal denoiser accurately, due to the peaked nature of $\pRcyx$. 
    }
    \label{fig:error-plot}
\end{figure}

In \autoref{fig:error-plot}, we compute error in the zeroth-order $\Dopt_0$, first-order $\Dopt_1$, and second-order $\Dopt_2$ approximations compared to an estimate of $\Dopt(y; x)$ obtained by numerically integrating the expectation in \autoref{eqn:optimal-denoiser-expectation} in Mathematica, which automatically adjusts the resolution of the grid for the numerical quadrature on $SO(3)$.

For larger $\sigma$, we see that the error rates drop significantly using higher-order correction terms. For smaller $\sigma$, we quickly reach the regime where error is dominated by numerical precision rather than approximation error when using higher-order correction terms. This plot confirms that the error in approximating the optimal conditional denoiser can be significantly reduced by our estimators.

In the next section, we experiment with the practical utility of these estimators.

% In the next section, we demonstrate the practical utility of these estimators in improving the training of diffusion models in two practical contexts: 1) Boltz-1 \citep{boltz1}, an open-source reproduction of the popular AlphaFold 3 protein folding model \citep{alphafold3}, and 2) a diffusion model for small peptides based on the NequIP \citep{nequip} architecture as demonstrated in JAMUN \citep{jamun}.

\section{Results in Practice}
\label{sec:practice}

\begin{figure}[h]
    \centering
    \includegraphics[width=0.7\textwidth]
    {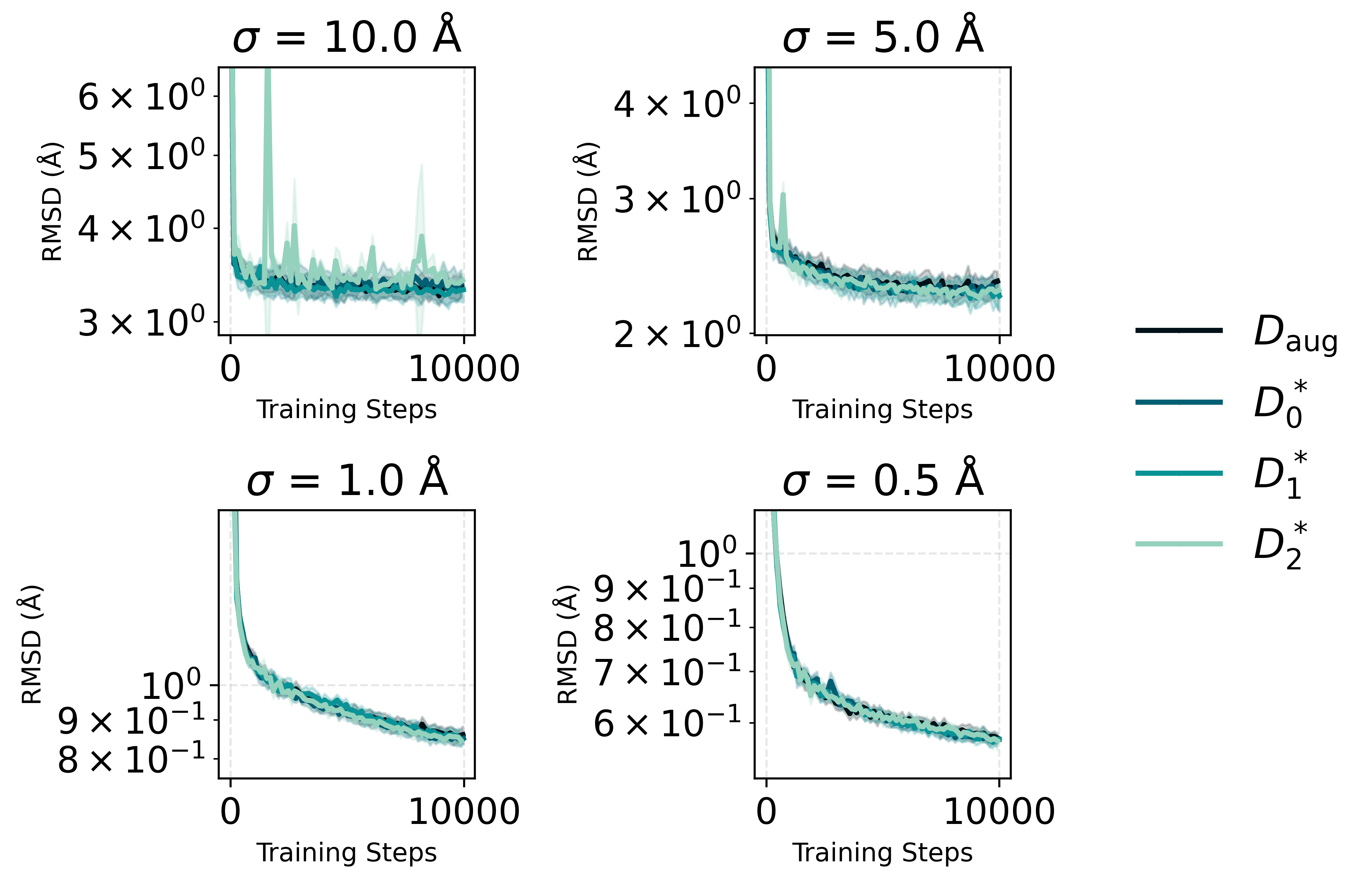}
    \caption{Training progress for the MLP, as measured by RMSD to ground-truth $x$ (sampled from all frames), when trained using $\Daug$, $\Dopt_0$, $\Dopt_1$ and $\Dopt_2$.}
    \label{fig:error-plot-mlp-rmsd-all-frames}
\end{figure}

\begin{figure}[h]
    \centering
    \includegraphics[width=0.7\textwidth]
    {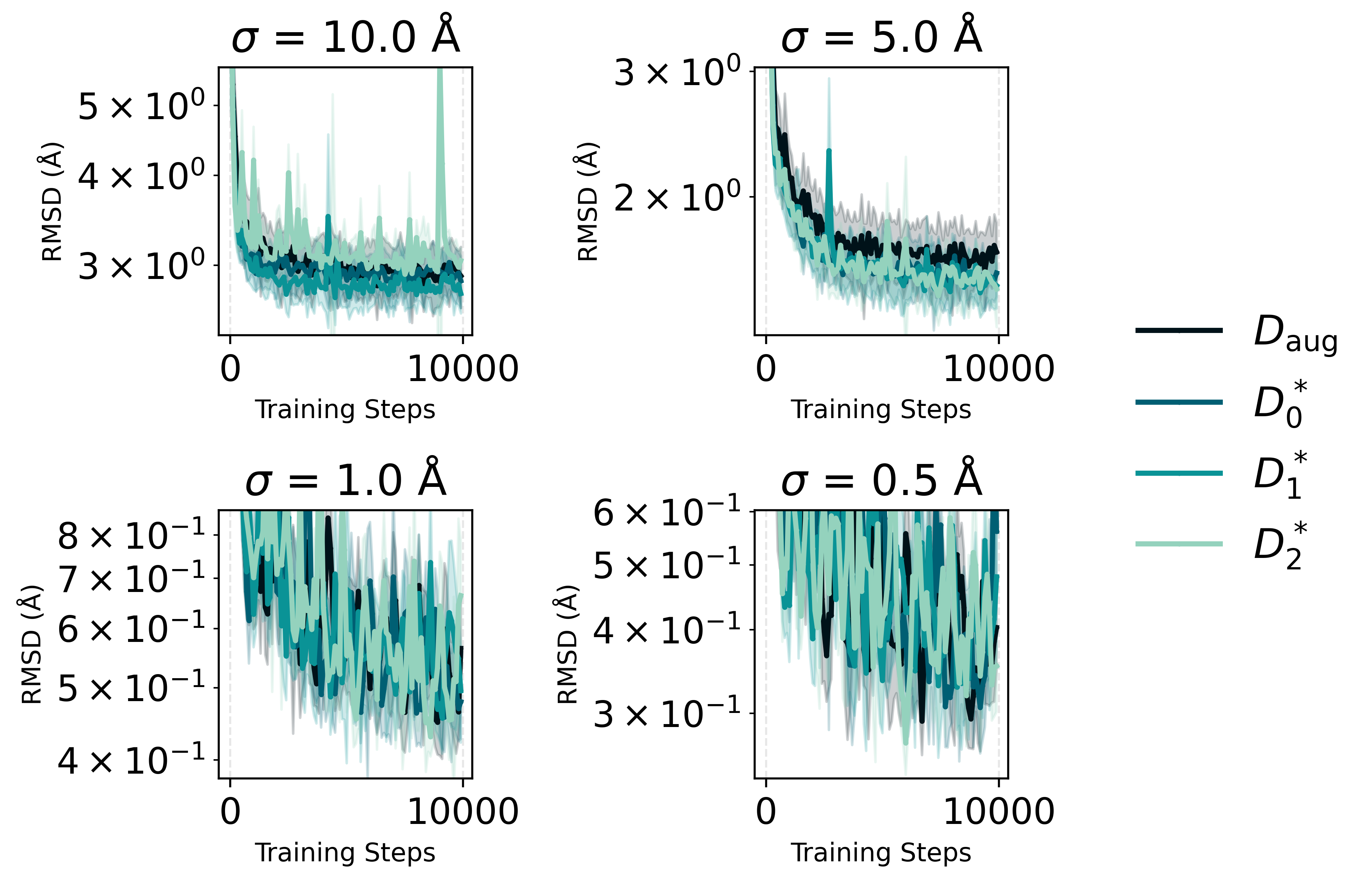}
    \caption{Training progress for the MLP, as measured by RMSD to ground-truth $x$ (fixed as the first frame), when trained using $\Daug$, $\Dopt_0$, $\Dopt_1$ and $\Dopt_2$. }
    \label{fig:error-plot-mlp-rmsd-single-frame}
\end{figure}

We train a simple $2$-layer MLP (with $2.3\si{\million}$ parameters) on 3D configurations of the tetrapeptide \texttt{AEQN} as obtained from the \textsc{Timewarp 4AA-Large} dataset \citep{timewarp}, utilizing the codebase of the JAMUN \citep{jamun} model. We train this model using the losses corresponding to the estimators $\Daug, \Dopt_0, \Dopt_1$ and $\Dopt_2$ discussed above, at $4$ different noise levels $\sigma = \SIlist{0.5;1.0;5;10}{\angstrom}$. This captures most of the noise levels usually used to train diffusion models on data of this size \citep{boltz1}. We perform two experiments which allow us to measure the impact of the estimators in learning 1. the optimal denoiser and 2. the optimal conditional denoiser.
\begin{enumerate}
    \item We sample $x$ from all $50000$ frames of a molecular dynamics simulation for the \texttt{AEQN} peptide, as obtained from \citet{timewarp}.
    \item We fix $x$ as the first frame of the same molecular dynamics simulation.
\end{enumerate}
The results are shown in \autoref{fig:error-plot-mlp-rmsd-all-frames} and \autoref{fig:error-plot-mlp-rmsd-single-frame}, where we report the RMSD (root mean square deviation) to the ground truth $x$. In \autoref{sec:additional-results}, we also show plots of the aligned RMSD for the same training runs.

We see that at the largest noise level, the second-order correction tends to diverge. At the lowest noise levels, the magnitude of the correction is not significant, and all estimators perform similarly. We see that in practice, the zeroth-order approximation $\Dopt_0$ is usually good enough at the important noise levels; in particular, it seems like the model is unable to take advantage of the variance reduction from the higher-order corrections. We hypothesize that this may be due to the fact that the variance of the gradients over the multiple $x$ is much greater than the variance of the gradients due to the rotational symmetries. These results are preliminary, but they suggest that alignment itself may not be super critical to learn a good denoiser. 

\clearpage

\bibliographystyle{bibstyle}
\bibliography{references}

\begin{thebibliography}{23}
\providecommand{\natexlab}[1]{#1}
\providecommand{\url}[1]{\texttt{#1}}
\expandafter\ifx\csname urlstyle\endcsname\relax
  \providecommand{\doi}[1]{doi: #1}\else
  \providecommand{\doi}{doi: \begingroup \urlstyle{rm}\Url}\fi

\bibitem[Abramson et~al.(2024)Abramson, Adler, Dunger, Evans, Green, Pritzel, Ronneberger, Willmore, Ballard, Bambrick, Bodenstein, Evans, Hung, O'Neill, Reiman, Tunyasuvunakool, Wu, {\v{Z}}emgulyt{\.{e}}, Arvaniti, Beattie, Bertolli, Bridgland, Cherepanov, Congreve, Cowen-Rivers, Cowie, Figurnov, Fuchs, Gladman, Jain, Khan, Low, Perlin, Potapenko, Savy, Singh, Stecula, Thillaisundaram, Tong, Yakneen, Zhong, Zielinski, {\v{Z}}{\'i}dek, Bapst, Kohli, Jaderberg, Hassabis, and Jumper]{alphafold3}
Josh Abramson, Jonas Adler, Jack Dunger, Richard Evans, Tim Green, Alexander Pritzel, Olaf Ronneberger, Lindsay Willmore, Andrew~J. Ballard, Joshua Bambrick, Sebastian~W. Bodenstein, David~A. Evans, Chia-Chun Hung, Michael O'Neill, David Reiman, Kathryn Tunyasuvunakool, Zachary Wu, Akvil{\.{e}} {\v{Z}}emgulyt{\.{e}}, Eirini Arvaniti, Charles Beattie, Ottavia Bertolli, Alex Bridgland, Alexey Cherepanov, Miles Congreve, Alexander~I. Cowen-Rivers, Andrew Cowie, Michael Figurnov, Fabian~B. Fuchs, Hannah Gladman, Rishub Jain, Yousuf~A. Khan, Caroline M.~R. Low, Kuba Perlin, Anna Potapenko, Pascal Savy, Sukhdeep Singh, Adrian Stecula, Ashok Thillaisundaram, Catherine Tong, Sergei Yakneen, Ellen~D. Zhong, Michal Zielinski, Augustin {\v{Z}}{\'i}dek, Victor Bapst, Pushmeet Kohli, Max Jaderberg, Demis Hassabis, and John~M. Jumper.
\newblock {Accurate structure prediction of biomolecular interactions with AlphaFold 3}.
\newblock \emph{Nature}, 630\penalty0 (8016):\penalty0 493--500, Jun 2024.
\newblock ISSN 1476-4687.
\newblock \doi{10.1038/s41586-024-07487-w}.
\newblock URL \url{https://doi.org/10.1038/s41586-024-07487-w}.

\bibitem[Campbell et~al.(2024)Campbell, Yim, Barzilay, Rainforth, and Jaakkola]{multiflow}
Andrew Campbell, Jason Yim, Regina Barzilay, Tom Rainforth, and Tommi Jaakkola.
\newblock Generative flows on discrete state-spaces: Enabling multimodal flows with applications to protein co-design, 2024.
\newblock URL \url{https://arxiv.org/abs/2402.04997}.

\bibitem[Daigavane et~al.(2025)Daigavane, Vani, Davidson, Saremi, Rackers, and Kleinhenz]{jamun}
Ameya Daigavane, Bodhi~P. Vani, Darcy Davidson, Saeed Saremi, Joshua Rackers, and Joseph Kleinhenz.
\newblock Jamun: Bridging smoothed molecular dynamics and score-based learning for conformational ensembles, 2025.
\newblock URL \url{https://arxiv.org/abs/2410.14621}.

\bibitem[Dunn \& Koes(2025)Dunn and Koes]{flowmol3}
Ian Dunn and David~R. Koes.
\newblock Flowmol3: Flow matching for 3d de novo small-molecule generation, 2025.
\newblock URL \url{https://arxiv.org/abs/2508.12629}.

\bibitem[Hoogeboom et~al.(2022)Hoogeboom, Satorras, Vignac, and Welling]{edm-molecules}
Emiel Hoogeboom, Victor~Garcia Satorras, Clément Vignac, and Max Welling.
\newblock Equivariant diffusion for molecule generation in 3d, 2022.
\newblock URL \url{https://arxiv.org/abs/2203.17003}.

\bibitem[Inc.()]{Mathematica}
Wolfram~Research{,} Inc.
\newblock Mathematica, {V}ersion 14.2.
\newblock URL \url{https://www.wolfram.com/mathematica}.
\newblock Champaign, IL, 2024.

\bibitem[Kabsch(1976)]{kabsch}
W.~Kabsch.
\newblock {A solution for the best rotation to relate two sets of vectors}.
\newblock \emph{Acta Crystallographica Section A}, 32\penalty0 (5):\penalty0 922--923, Sep 1976.
\newblock \doi{10.1107/S0567739476001873}.
\newblock URL \url{https://doi.org/10.1107/S0567739476001873}.

\bibitem[Kadkhodaie et~al.(2024)Kadkhodaie, Guth, Simoncelli, and Mallat]{kadkhodaie2024generalization}
Zahra Kadkhodaie, Florentin Guth, Eero~P Simoncelli, and St{\'e}phane Mallat.
\newblock Generalization in diffusion models arises from geometry-adaptive harmonic representations.
\newblock In \emph{The Twelfth International Conference on Learning Representations}, 2024.
\newblock URL \url{https://openreview.net/forum?id=ANvmVS2Yr0}.

\bibitem[Karras et~al.(2022)Karras, Aittala, Aila, and Laine]{edm}
Tero Karras, Miika Aittala, Timo Aila, and Samuli Laine.
\newblock Elucidating the design space of diffusion-based generative models, 2022.

\bibitem[Klein et~al.(2023{\natexlab{a}})Klein, Foong, Fjelde, Mlodozeniec, Brockschmidt, Nowozin, Noé, and Tomioka]{timewarp}
Leon Klein, Andrew Y.~K. Foong, Tor~Erlend Fjelde, Bruno Mlodozeniec, Marc Brockschmidt, Sebastian Nowozin, Frank Noé, and Ryota Tomioka.
\newblock Timewarp: Transferable acceleration of molecular dynamics by learning time-coarsened dynamics, 2023{\natexlab{a}}.
\newblock URL \url{https://arxiv.org/abs/2302.01170}.

\bibitem[Klein et~al.(2023{\natexlab{b}})Klein, Kr{\"a}mer, and Noe]{klein2023equivariant}
Leon Klein, Andreas Kr{\"a}mer, and Frank Noe.
\newblock Equivariant flow matching.
\newblock In \emph{Thirty-seventh Conference on Neural Information Processing Systems}, 2023{\natexlab{b}}.
\newblock URL \url{https://openreview.net/forum?id=eLH2NFOO1B}.

\bibitem[Lee(2018)]{lee2018bayesian}
Taeyoung Lee.
\newblock {Bayesian Attitude Estimation With the Matrix Fisher Distribution on SO(3)}.
\newblock \emph{IEEE Transactions on Automatic Control}, 63\penalty0 (10):\penalty0 3377--3392, 2018.
\newblock \doi{10.1109/TAC.2018.2797162}.

\bibitem[Li et~al.(2023)Li, Li, Zhang, and Bian]{li2023on}
Puheng Li, Zhong Li, Huishuai Zhang, and Jiang Bian.
\newblock On the generalization properties of diffusion models.
\newblock In \emph{Thirty-seventh Conference on Neural Information Processing Systems}, 2023.
\newblock URL \url{https://openreview.net/forum?id=hCUG1MCFk5}.

\bibitem[Mohlin et~al.(2020)Mohlin, Sullivan, and Bianchi]{mohlin2020probabilistic}
David Mohlin, Josephine Sullivan, and G{\'e}rald Bianchi.
\newblock Probabilistic orientation estimation with matrix fisher distributions.
\newblock \emph{Advances in Neural Information Processing Systems}, 33:\penalty0 4884--4893, 2020.

\bibitem[Niedoba et~al.(2024)Niedoba, Green, Naderiparizi, Lioutas, Lavington, Liang, Liu, Zhang, Dabiri, Ścibior, Zwartsenberg, and Wood]{niedoba2024nearestneighbourscoreestimators}
Matthew Niedoba, Dylan Green, Saeid Naderiparizi, Vasileios Lioutas, Jonathan~Wilder Lavington, Xiaoxuan Liang, Yunpeng Liu, Ke~Zhang, Setareh Dabiri, Adam Ścibior, Berend Zwartsenberg, and Frank Wood.
\newblock Nearest neighbour score estimators for diffusion generative models, 2024.
\newblock URL \url{https://arxiv.org/abs/2402.08018}.

\bibitem[Song et~al.(2022)Song, Meng, and Ermon]{ddim}
Jiaming Song, Chenlin Meng, and Stefano Ermon.
\newblock Denoising diffusion implicit models, 2022.
\newblock URL \url{https://arxiv.org/abs/2010.02502}.

\bibitem[Umeyama(1991)]{umeyama}
S.~Umeyama.
\newblock Least-squares estimation of transformation parameters between two point patterns.
\newblock \emph{IEEE Transactions on Pattern Analysis and Machine Intelligence}, 13\penalty0 (4):\penalty0 376--380, 1991.
\newblock \doi{10.1109/34.88573}.

\bibitem[Vastola(2025)]{vastola}
John~J. Vastola.
\newblock Generalization through variance: how noise shapes inductive biases in diffusion models, 2025.
\newblock URL \url{https://arxiv.org/abs/2504.12532}.

\bibitem[Wohlwend et~al.(2024)Wohlwend, Corso, Passaro, Reveiz, Leidal, Swiderski, Portnoi, Chinn, Silterra, Jaakkola, and Barzilay]{boltz1}
Jeremy Wohlwend, Gabriele Corso, Saro Passaro, Mateo Reveiz, Ken Leidal, Wojtek Swiderski, Tally Portnoi, Itamar Chinn, Jacob Silterra, Tommi Jaakkola, and Regina Barzilay.
\newblock Boltz-1 democratizing biomolecular interaction modeling.
\newblock \emph{bioRxiv}, 2024.
\newblock \doi{10.1101/2024.11.19.624167}.
\newblock URL \url{https://www.biorxiv.org/content/early/2024/11/20/2024.11.19.624167}.

\bibitem[Xu et~al.(2022)Xu, Yu, Song, Shi, Ermon, and Tang]{geodiff}
Minkai Xu, Lantao Yu, Yang Song, Chence Shi, Stefano Ermon, and Jian Tang.
\newblock {GeoDiff: a Geometric Diffusion Model for Molecular Conformation Generation}, 2022.
\newblock URL \url{https://arxiv.org/abs/2203.02923}.

\bibitem[Yang et~al.(2024)Yang, Chen, Wang, Liu, and Chen]{ddsm}
Shuai Yang, Yukang Chen, Luozhou Wang, Shu Liu, and Yingcong Chen.
\newblock Denoising diffusion step-aware models, 2024.
\newblock URL \url{https://arxiv.org/abs/2310.03337}.

\bibitem[Yim et~al.(2023)Yim, Campbell, Foong, Gastegger, Jim{\'e}nez-Luna, Lewis, Satorras, Veeling, Barzilay, Jaakkola, et~al.]{yim2023fast}
Jason Yim, Andrew Campbell, Andrew~YK Foong, Michael Gastegger, Jos{\'e} Jim{\'e}nez-Luna, Sarah Lewis, Victor~Garcia Satorras, Bastiaan~S Veeling, Regina Barzilay, Tommi Jaakkola, et~al.
\newblock Fast protein backbone generation with se (3) flow matching.
\newblock \emph{arXiv preprint arXiv:2310.05297}, 2023.

\bibitem[Yim et~al.(2024)Yim, Campbell, Mathieu, Foong, Gastegger, Jimenez-Luna, Lewis, Satorras, Veeling, Noe, Barzilay, and Jaakkola]{yim2024improved}
Jason Yim, Andrew Campbell, Emile Mathieu, Andrew Y.~K. Foong, Michael Gastegger, Jose Jimenez-Luna, Sarah Lewis, Victor~Garcia Satorras, Bastiaan~S. Veeling, Frank Noe, Regina Barzilay, and Tommi Jaakkola.
\newblock Improved motif-scaffolding with {SE}(3) flow matching.
\newblock \emph{Transactions on Machine Learning Research}, 2024.
\newblock ISSN 2835-8856.
\newblock URL \url{https://openreview.net/forum?id=fa1ne8xDGn}.

\end{thebibliography}

\clearpage

\appendix

\section{Proofs}
\label{sec:proofs}

\subsection{Equivariant Denoisers Sample Invariant Distributions}
\label{sec:equivariant-learns-invariant}

The proof is similar to that of Proposition 1 in \citet{geodiff}.

Suppose that $p_{y}(; \sigma_{i + 1})$ is a $SO(3)$-invariant distribution:
\begin{align}
    p_{y}(y_{i + 1}; \sigma_{i + 1}) = p_{y}(\R \circ y_{i + 1}; \sigma_{i + 1})
\end{align}

Using the DDIM update rule (\autoref{eqn:ddim}) at step $(i + 1)$, we see that the next distribution $p_{y}(; \sigma_{i})$ is also $SO(3)$-invariant.
Under an arbitrary rotation $\R$:
\begin{align}
    \label{eqn:ddim-repeat}
    \operatorname{DDIM}_{i + 1}(\R \circ y_{i + 1}) &= \R \circ y_{i + 1} + \left(1 - \frac{\sigma_i}{\sigma_{i + 1}}\right)(D(\R \circ  y_{i + 1}, \sigma_{i + 1}) - \R \circ y_{i + 1})
    \\
    &= \R \circ y_{i + 1} + \left(1 - \frac{\sigma_i}{\sigma_{i + 1}}\right)(\R \circ D(y_{i + 1}, \sigma_{i + 1}) - \R \circ y_{i + 1})
    \\
    &= \R \circ \left(y_{i + 1} + \left(1 - \frac{\sigma_i}{\sigma_{i + 1}}\right)(D(y_{i + 1}, \sigma_{i + 1}) - y_{i + 1})\right)
    \\
    &= \R \circ y_{i + 1}
    \\
    &= \R \circ \operatorname{DDIM}_{i + 1}(y_{i + 1})
\end{align}
We used the $SO(3)$-equivariance of the denoiser above: $D(\R \circ  y_{i + 1}, \sigma_{i + 1}) = \R \circ D(y_{i + 1}, \sigma_{i + 1})$ for all $y_{i + 1}$.

Hence:
\begin{align}
    p_{y}(y_{i}; \sigma_{i}) &= \int p(y_i | y_{i + 1})  p_{y}(y_{i + 1}; \sigma_{i + 1}) dy_{i + 1} \\
    &= \int \delta(y_{i} - \operatorname{DDIM}_{i + 1}(y_{i + 1})) p_{y}(y_{i + 1}; \sigma_{i + 1})  dy_{i + 1} \\
    &= \int \delta(y_{i} - \operatorname{DDIM}_{i + 1}(y_{i + 1})) p_{y}(\R \circ y_{i + 1}; \sigma_{i + 1})  dy_{i + 1} \\
    &= \int \delta(\R \circ y_{i} - \R \circ \operatorname{DDIM}_{i + 1}(y_{i + 1})) p_{y}(\R \circ y_{i + 1}; \sigma_{i + 1})  dy_{i + 1} \\
    &= \int \delta(\R \circ y_{i} - \operatorname{DDIM}_{i + 1}(\R \circ y_{i + 1})) p_{y}(\R \circ y_{i + 1}; \sigma_{i + 1})  dy_{i + 1} \\
    &= \int \delta(\R \circ y_{i} - \operatorname{DDIM}_{i + 1}(y_{i + 1}')) p_{y}(y_{i + 1}'; \sigma_{i + 1})  dy_{i + 1}' \\
    &= p_{y}(\R \circ y_{i}; \sigma_{i})
\end{align}
using the change of variables $y_{i + 1}' \equiv \R \circ y_{i + 1}$ which does not induce any change of measure.
Hence, $$p_{y}(; \sigma_{i})$$ is a $SO(3)$-invariant distribution. The initial distribution is $\N(0, \sigma_M^2 \II_d)$, which is also $SO(3)$-invariant distribution due to the isotropy of the multivariate Gaussian. We can conclude that at each noise level $\sigma_i$ in the diffusion process, the $p_{y}(; \sigma_{i})$ is $SO(3)$-invariant.

The same proof can be generalized to stochastic samplers such as DDSM \citep{ddsm}.

\subsection{Rotationally Augmented Distributions are Invariant}
\label{sec:augmentation-makes-invariant}
For any arbitrary rotation $\R$:
\begin{align}
    \aug[p_x](\R \circ x)= & \int_{SO(3)}p_x(\R'^{-1} \R \circ x)\unifR(\R')\mathrm{d}\R' \\
    = & \int_{SO(3)}p_x((\R^{-1}\R')^{-1} \circ x)\unifR(\R(\R^{-1}\R'))\mathrm{d}(\R\R^{-1}\R') \\
    = & \int_{SO(3)}p_x(\R''^{-1} \circ x)\unifR(\R\R'')\mathrm{d}(\R\R'') \\
    = & \int_{SO(3)}p_x(\R''^{-1} \circ x)\unifR(\R'')\mathrm{d}\R'' \\
    = & \aug[p_x](x).
\end{align}

\subsection{Rotational Augmentation Does Not Affect Equivariant Denoisers}
\label{sec:augmentation-equivariant}

Using the equivariance of $D$ and the fact that for any rotation $\R$, we have $\R^T\R = \II_{3 \times 3}$:
\begin{align}
    \norm{D(\R(x + \eta)) - \R x}^2
    &= \norm{\R D(x + \eta) - \R x}^2  \\
    &= \norm{\R (D(x + \eta) - x)}^2  \\
    &= (\R (D(x + \eta) - x))^T{\R (D(x + \eta) - x)}  \\
    &= (D(x + \eta) - x))^T \R^T\R (D(x + \eta) - x)  \\
    &= (D(x + \eta) - x))^T (D(x + \eta) - x)  \\
    &= \norm{(D(x + \eta) - x)}^2 
\end{align}
Hence, the loss $\lnoaug(D)$ is invariant under rotations $\R$.
Thus, if $D$ is equivariant:
\begin{align}
    \lnoaug(D) =  \EE_{\R \sim \unifR} \EE_{x \sim p_x}  \EE_{\eta \sim p_\eta} &\norm{D(\R (x + \eta)) - \R x}^2
    = \laug(D)
\end{align}

\section{No Perfect Denoiser Exists with Rotational Augmentation}
\label{sec:no-perfect-denoiser}
Suppose we only had one sample $\xz$ where $\|\xz\|>0$. Hence, our distribution is a delta function $p_x(x) = \delta(x - \xz)$. Fix a noise level $\sigma>0$. Here, we argue why a perfect denoiser cannot exist\footnote{This argument can be made rigorous to allow for exceptions of zero measure. We do this in \autoref{sec:no-perfect-denoiser-rigorous}, but ignore such exceptions for clarity here.}.

In this setting, assume that there exists a perfect denoiser $\Dperf$:
\begin{align}
    \label{eqn:perfect}
    \Dperf(\R \circ (\xz + \sigma\eta)) = \R \circ \xz
\end{align}
for all rotations $\R$ and noise $\eta$.
Such a $\Dperf$ obtains zero loss: $\laug(\Dperf) = 0$.

We show that such a $\Dperf$ cannot exist, by contradiction. 
Set $\R = \II$ in \autoref{eqn:perfect}, to see that $\Dperf(\xz + \sigma\eta) = \xz$ for all instantiations of $\eta$. Fix some noise $\eta_0$ arbitrarily. Now, consider any non-identity rotation $\R$. Let $\eta_\R$ be such that:
\begin{align}
    \R \circ (\xz + \sigma \eta_\R) = \xz + \sigma \eta_0 
    \quad \implies \quad \eta_\R = \frac{\R^\top \circ (\xz + \sigma \eta_0) - \xz}{\sigma}
\end{align}
where we used the fact that $\R$ is orthonormal.
Now, setting $\eta = \eta_\R$ in \autoref{eqn:perfect}:
\begin{align}
    \Dperf(\R \circ (\xz + \sigma\eta_\R)) = \R \circ \xz.
\end{align} But from the definition of $\eta_\R$, we have:
\begin{align}
    \Dperf(\R \circ (\xz + \sigma \eta_\R)) = \Dperf(\xz + \sigma\eta_0) = \xz.
\end{align}
As $\R$ is not the identity, we have a contradiction. Thus, $\Dperf$ cannot exist.

\subsection{No Perfect Denoiser Exists with Rotational Augmentation}
\label{sec:no-perfect-denoiser-rigorous}
We can formalize the argument of \autoref{sec:no-perfect-denoiser} to include exceptions of measure $0$.

Assume that there exists a perfect denoiser $\Dperf$:
\begin{align}
    \Dperf(\R \circ (\xz + \sigma\eta)) = \R \circ \xz
\end{align}
for all rotations $\R$ and noise $\eta$, except on a set $S$ of measure $0$ over $SO(3) \times \RR^{N \times 3}$.
Now, define the family of sets:
\begin{align}
    R(y) = \{ (\R, \eta) : \R \circ (\xz + \sigma\eta) = y \}
\end{align}
for each $y \in \RR^{N \times 3}$.
It is easy to verify that $\{R(y)\}_{y \in \RR^{N \times 3}}$ is a partition of the space $SO(3) \times \RR^{N \times 3}$. Further, each $R(y)$ is diffeomorphic to $SO(3)$, since we can always find a $\eta$ for every $\R$ to satisfy $\R \circ (\xz + \sigma\eta) = y$.
Thus, we have $\mu_{SO(3)}(R(y)) = 1$ where $\mu_{SO(3)}$ is the Haar measure over $SO(3)$.

Thus, we can measure any measurable set $A$ by integrating the measure of its intersections with each $R(y)$.
Formally, by the co-area formula, where $\mu$ represents the product measure over $SO(3) \times \RR^{N \times 3}$:
\begin{align}
    \mu(A) = \int_{y \in \RR^{N \times 3}} \mu_{SO(3)}(A \cap R(y)) J(y) d y
\end{align}
where $J(y) > 0$ is the Jacobian factor. (Essentially, this is the `change of variables' formula.)
Now, applying this to $S$ with $\mu(S) = 0$, we see that:
\begin{align}
    \int_{y \in \RR^{N \times 3}} \mu_{SO(3)}(S \cap R(y)) J(y) d y &= 0 \\
    \implies \mu_{SO(3)}(S \cap R(y)) &= 0 \ \text{for almost every} \ y \in \RR^{N \times 3}
\end{align}
Pick any such $y$ . Then, 
\begin{align}
    \mu_{SO(3)}(R(y) - S) &= \mu_{SO(3)}(R(y))  - \mu_{SO(3)}(S \cap R(y)) = 1 - 0 = 1.
\end{align}
Thus, for this particular $y$, $\mu_{SO(3)}(R(y) - S) = 1 > 0 \implies R(y) - S$ is uncountable, and has infinitely many points. In particular, there exist atleast two different points $(\R_1, \eta_1)$ and $(\R_2, \eta_2)$ in $R(y) - S$ such that $\R_1 \neq \R_2$.
Now, for each of these two points:
\begin{align}
    \R_1 \circ (\xz + \sigma\eta_1) = y = \R_2 \circ (\xz + \sigma\eta_2). 
\end{align}
But, since $(\R_1, \eta_1)$ and $(\R_2, \eta_2)$ in $R(y) - S$ (and hence, not in $S$), we must have perfect denoising for the corresponding inputs:
\begin{align}
    \R_1 \circ \xz  = \Dperf(\R_1 \circ (\xz + \sigma\eta_1)) = \Dperf(y) =  \Dperf(\R_2 \circ (\xz + \sigma\eta_2)) = \R_2 \circ \xz
\end{align}
which is a contradiction as $\|\xz\| > 0$ and $\R_1 \neq \R_2$. Thus, perfect denoising is impossible.

\subsection{The Optimal Denoiser in the Single Sample Setting}
\label{sec:proof-single-sample-optimal}
In the single sample case where $p_x(x) = \delta(x - x_0)$, using the isotropy of the Gaussian $p_\eta$:
\begin{align}
    \laug(D) &= \EE_{\R \sim \unifR} \EE_{\eta \sim \N(0, \sigma^2 \II_{N \times 3})} \norm{D(\R (x_0 + \eta)) - \R x_0}^2  \\
    &= \EE_{\R \sim \unifR} \EE_{\eta \sim \N(0, \sigma^2 \II_{N \times 3})} \norm{D(\R x_0 + \eta) - \R x_0}^2  \\
    &= \EE_{\R \sim \unifR} \EE_{y \sim \N(\R x_0, \sigma^2 \II_{N \times 3})} \norm{D(y) - \R x_0}^2  \\
    &= \int_{SO(3)} \left(\int_{\RR^{N \times 3}} \norm{D(y) - \R x_0}^2 \N(y; \R x_0, \sigma^2 \II_{N \times 3}) dy \right) \unifR(\R) d\R  \\
    &=  \int_{\RR^{N \times 3}} \left(\int_{SO(3)} \norm{D(y) - \R x_0}^2 \N(y; \R x_0, \sigma^2 \II_{N \times 3}) \unifR(\R) d\R \right)  dy   \\
    &= \int_{\RR^{N \times 3}} \laug(D; y, \sigma)dy
\end{align}
where we define:
\begin{align}
    \label{eqn:led-single}
    \laug(D; y, \sigma)&= \int_{SO(3)} \norm{D(y) - \R x_0}^2 \N(y; \R x_0, \sigma^2 \II_{N \times 3}) \unifR(\R) d\R  \\
    &= \EE_{\R \sim \unifR} \norm{D(y) - \R x_0}^2 \N(y; \R x_0, \sigma^2 \II_{N \times 3})
\end{align}
Note that $\laug(D; y)$ is non-negative for all $y$.
Thus, the optimal denoiser $\Dopt$ should minimize $\laug(D; y)$ for each possible $y$. Taking the gradient of $\laug(D; y)$ with respect to $D(y)$ and setting it to $0$:
\begin{align}
    \label{eqn:grad_le}
    &\nabla_{D(y)} \laug(D; y, \sigma)= 0  \\ 
    &\implies \EE_{\R \sim \unifR} \left[2(\Dopt(y) - \R x_0) \N(y; \R x_0, \sigma^2 \II_{N \times 3}) \right] = 0  \\
    &\implies \Dopt(y) = \frac{\EE_{\R \sim \unifR} [\R x_0 \ \N(y; \R x_0, \sigma^2 \II_{N \times 3})]}{\EE_{\R \sim \unifR} [\N(y; \R x_0, \sigma^2 \II_{N \times 3})]}
\end{align}
This is the optimal denoiser in the single sample setting!
We can rewrite this a bit:
\begin{align}
    \label{eqn:single-sample-optimal}
    \Dopt(y) &= \frac{\int_{SO(3)} \R x_0 \N(y; \R x_0, \sigma^2 \II_{N \times 3}) \unifR(\R) d\R}{\int_{SO(3)} \N(y; \R x_0, \sigma^2 \II_{N \times 3}) \unifR(\R) d\R}  \\
    &= \frac{\int_{SO(3)} \R x_0 \N(y; \R x_0, \sigma^2 \II_{N \times 3}) \unifR(\R) d\R}{\int_{SO(3)} \N(y; \R' x_0, \sigma^2 \II_{N \times 3}) \unifR(\R') d\R'}  \\
    &= \int_{SO(3)} \R x_0 \frac{\N(y; \R x_0, \sigma^2 \II_{N \times 3}) \unifR(\R) d\R}{\int_{SO(3)} \N(y; \R' x_0, \sigma^2 \II_{N \times 3}) \unifR(\R') d\R'}  \\
    &= \int_{SO(3)} \R x_0 \ p(\R \ | \ y, x_0)d\R  \\
    &= \EE_{\R \sim p(\R \ | \ y, x_0)} [\R x_0]
\end{align}
as:
\begin{align}
\label{eqn:induced-distribution-R-proof}
    p(\R \ | \ y, x_0) = \frac{\N(y; \R x_0, \sigma^2 \II_{N \times 3}) \unifR(\R)}{\int_{SO(3)} \N(y; \R' x_0, \sigma^2 \II_{N \times 3}) \unifR(\R') d\R'}
\end{align}
is the distribution over rotations $\R$ conditional on $y$, because:
\begin{align}
    p(y \ | \ \R, x_0) = \N(y; \R x_0, \sigma^2 \II_{N \times 3})
\end{align}
and Bayes' rule:
\begin{align}
    p(\R \ | \ y, x_0) = \frac{p(y, \R \ | \ x_0)}{p(y \ | \ x_0)} = \frac{p(y \ | \ \R, x_0) p(\R \ | \ x_0)}{\int_{SO(3)} p(y \ | \ \R', x_0) p(\R' \ | \ x_0) d\R'}
\end{align}
and noticing that $p(\R \ | \ x_0) = \unifR(\R)$.

\subsection{The Optimal Denoiser in the General Setting}
\label{sec:proof-multi-sample-optimal}
The same idea and calculations from \autoref{sec:single-sample-optimal} hold in the general setting, where $p_x$ is arbitrary.

Using the isotropy of the Gaussian $p_\eta$:
\begin{align}
    \laug(D) &= \EE_{\R \sim \unifR}\EE_{x \sim p_x} \EE_{\eta \sim \N(0, \sigma^2 \II_{N \times 3})} \norm{D(\R (x + \eta)) - \R \circ x}^2  \\
    &= \EE_{\R \sim \unifR} \EE_{x \sim p_x} \EE_{\eta \sim \N(0, \sigma^2 \II_{N \times 3})} \norm{D(\R \circ x + \eta) - \R \circ x}^2  \\
    &= \EE_{\R \sim \unifR} \EE_{x \sim p_x} \EE_{y \sim \N(\R \circ x, \sigma^2 \II_{N \times 3})} \norm{D(y) - \R \circ x}^2  \\
    &= \int_{SO(3)} \int_{\RR^{N \times 3}} \left(\int_{\RR^{N \times 3}} \norm{D(y) - \R \circ x_0}^2 \N(y; \R \circ x, \sigma^2 \II_{N \times 3}) dy \right) p_x(x) dx \ \unifR(\R) d\R  \\
    &=  \int_{\RR^{N \times 3}} \left(\int_{\RR^{N \times 3}} \left(\int_{SO(3)} \norm{D(y) - \R \circ x}^2 \N(y; \R \circ x, \sigma^2 \II_{N \times 3}) \unifR(\R) d\R \right)  \ p_x(x) dx \right) dy   \\
    &= \int_{\RR^{N \times 3}} \laug(D; y, \sigma)dy
\end{align}
where:
\begin{align}
    \label{eqn:led}
    \laug(D; y, \sigma)&= \int_{\RR^{N \times 3}} \int_{SO(3)} \norm{D(y) - \R \circ x}^2 \N(y; \R \circ x, \sigma^2 \II_{N \times 3}) \ \unifR(\R) d\R \ p_x(x) \ dx    \\
    &= \EE_{x \sim p_x} \EE_{\R \sim \unifR} \norm{D(y) - \R \circ x}^2 \N(y; \R \circ x, \sigma^2 \II_{N \times 3})
\end{align}
Note that $\laug(D; y)$ is non-negative for all $y$.
Thus, the optimal denoiser $\Dopt$ should minimize $\laug(D; y)$ for each possible $y$. Taking the gradient of $\laug(D; y)$ with respect to $D(y)$ and setting it to $0$:
\begin{align}
 \label{eqn:multi-sample-optimal}
    &\nabla_{D(y)} \laug(D; y, \sigma)= 0  \\ 
    &\implies \EE_{x \sim p_x} \EE_{\R \sim \unifR} \left[2(\Dopt(y) - \R \circ x) \N(y; \R \circ x, \sigma^2 \II_{N \times 3}) \right] = 0  \\
    &\implies \Dopt(y) = \frac{\EE_{x \sim p_x}\EE_{\R \sim \unifR}[\R \circ x \ \N(y; \R \circ x, \sigma^2 \II_{N \times 3})]}{\EE_{x \sim p_x}\EE_{\R \sim \unifR} [\N(y; \R \circ x, \sigma^2 \II_{N \times 3})]}
\end{align}
which clearly specializes to \autoref{eqn:single-sample-optimal} in the single sample setting. 
As before, we can write this as:
\begin{align}
     \label{eqn:multi-sample-optimal-simplified}
    \Dopt(y) &= \frac{\EE_{x \sim p_x}\EE_{\R \sim \unifR} \R \circ x \ \N(y; \R \circ x, \sigma^2 \II_{N \times 3})}{\EE_{x \sim p_x}\EE_{\R \sim \unifR} \N(y; \R \circ x, \sigma^2 \II_{N \times 3})}
     \\
    &=  \frac{\int_{SO(3)} \int_{\RR^{N \times 3}} \R \circ x \ \N(y; \R \circ x, \sigma^2 \II_{N \times 3}) p_x(x) \unifR(\R) dx d\R}{\int_{SO(3)} \int_{\RR^{N \times 3}} \N(y; \R' x', \sigma^2 \II_{N \times 3}) p_x(x') \unifR(\R') dx' d\R'}  \\
    &=  \int_{SO(3)} \int_{\RR^{N \times 3}} \R \circ x \ \frac{\N(y; \R \circ x, \sigma^2 \II_{N \times 3}) p_x(x) \unifR(\R) dx d\R}{\int_{SO(3)} \int_{\RR^{N \times 3}} \N(y; \R' x', \sigma^2 \II_{N \times 3}) p_x(x') \unifR(\R') dx' d\R'}  \\
    &=  \int_{SO(3)} \int_{\RR^{N \times 3}} \R \circ x \ \pxRcy dx d\R  \\
    &=  \EE_{x, \R \sim \pxRcy} [\R \circ x] \\
    &=  \EE_{x \sim \pxcy}[\EE_{\R \sim \pRcyx} [\R \circ x]] \\
    &=  \EE_{x \sim \pxcy}[\Dopt(y; x)]
\end{align}
as the conditional probability distribution $\pxRcy$ over both point clouds $x$ and rotations $\R$ is:
\begin{align}
    \label{eqn:induced-distribution-joint}
    \pxRcy &= \frac{\pycRx p(x)p(\R)}{\int_x \int_\R \pycRx p(x)p(\R)} \\
    &= \frac{\N(y; \R \circ x, \sigma^2 \II_{N \times 3}) p_x(x) \unifR(\R)}{\int_{SO(3)} \int_{\RR^{N \times 3}} \N(y; \R' x', \sigma^2 \II_{N \times 3}) p_x(x') \unifR(\R') dx' d\R'}
\end{align}
Note that the marginal distribution over $\R$ under $\pxRcy$ is indeed $\pRcyx$ as derived in \autoref{eqn:induced-distribution-R}.
Note that:
\begin{align}
\label{eqn:bayes-induced-distribution-R}
    \pRcyx = \frac{\N(y; \R \circ x, \sigma^2 \II_{N \times 3}) \unifR(\R)}{\int_{SO(3)} \N(y; \R' x, \sigma^2 \II_{N \times 3}) \unifR(\R') d\R'}
\end{align}
because:
\begin{align}
    \pycRx = \N(y; \R \circ x, \sigma^2 \II_{N \times 3})
\end{align}
and Bayes' rule:
\begin{align}
    p(\R \ | \ y, x) = \frac{p(y, \R \ | \ x)}{p(y \ | \ x)} = \frac{\pycRx p(\R \ | \ x)}{\int_{SO(3)} p(y \ | \ \R', x) p(\R' \ | \ x) d\R'}
\end{align}
and noticing that $p(\R \ | \ x) = \unifR(\R)$.

\subsection{Connection to the Matrix Fisher Distribution}
\label{sec:matrix-fisher-connection}

Here, we show that $p(\R \ | \ y; x_0, \sigma)$ belongs to the family of Matrix Fisher distributions:
\begin{align}
    p(\R \ | \ y; x_0, \sigma) &\propto \N(y; \R \circ x_0, \sigma^2 \II_{N \times 3}) \unifR(\R) \\
    &\propto \exp\left(-\frac{\norm{y - \R \circ x_0}^2}{2\sigma^2}\right)
    \\
    &\propto \exp\left(-\frac{\norm{y}^2 + 
    \norm{\R \circ x_0}^2 - 2\Tr[y^T (\R \circ x_0)] }{2\sigma^2}\right)
    \\
    &\propto \exp\left(-\frac{\norm{y}^2 + 
    \norm{x_0}^2 - 2\Tr[y^T (\R \circ x_0)]}{2\sigma^2}\right)
    \\
    &\propto \exp\left(\frac{\Tr[y^T (\R \circ x_0)]}{\sigma^2}\right)
    \\
    &\propto \exp\left(\frac{\Tr[y^T (x_0\R^T)]}{\sigma^2}\right)
    \\
    &\propto \exp\left(\frac{\Tr[\R^T y^T x_0]}{\sigma^2}\right)
    \\
    &\propto \exp\left(\frac{\Tr[(y^T x_0)^T \R]}{\sigma^2}\right)
    \\
    &\propto \exp\left(\Tr\left[ \left(\frac{y^T x_0}{\sigma^2}\right)^T\R\right]\right)
\end{align}
Hence, $p(\R \ | \ y; x_0, \sigma) = \MF(\R; \frac{y^T x_0}{\sigma^2})$.

% \subsubsection{The Optimal Denoiser is $SO(3)$-Equivariant}
% \label{sec:optimal-denoiser-equivariant-general}
% The optimal denoiser is also \emph{equivariant} in the general setting:
% \begin{align}
%     \Dopt(\R' y) &= 
%     \frac{\EE_{x \sim p_x}\EE_{\R \sim \unifR} [\R \circ x \ \N(\R'y; \R \circ x, \sigma^2 \II_{N \times 3})]}{\EE_{x \sim p_x}\EE_{\R \sim \unifR} [\N(\R'y; \R \circ x, \sigma^2 \II_{N \times 3})]} 
%      \\
%     &= 
%     \frac{\EE_{x \sim p_x}\EE_{\R \sim \unifR} [\R \circ x \ \N(y; (\R')^{-1}\R \circ x, \sigma^2 \II_{N \times 3})]}{\EE_{x \sim p_x}\EE_{\R \sim \unifR} [\N(y; (\R')^{-1}\R \circ x, \sigma^2 \II_{N \times 3})]} 
%      \\
%     &= 
%     \frac{\EE_{x \sim p_x}\EE_{\R \sim \unifR} [\R'((\R')^{-1}\R) x \ \N(y; ((\R')^{-1}\R) x, \sigma^2 \II_{N \times 3})]}{\EE_{x \sim p_x}\EE_{\R \sim \unifR} [\N(y; ((\R')^{-1}\R) x, \sigma^2 \II_{N \times 3})]} 
%      \\
%     &= \frac{\EE_{x \sim p_x}\EE_{\R'' \sim \unifR} [\R' \R'' x \ \N(y; \R'' x, \sigma^2 \II_{N \times 3})]}{\EE_{x \sim p_x}\EE_{\R'' \sim \unifR} [\N(y; \R'' x, \sigma^2 \II_{N \times 3})]} 
%      \\
%     &= \R' \frac{\EE_{x \sim p_x}\EE_{\R'' \sim \unifR} [ \R'' x \ \N(y; \R'' x, \sigma^2 \II_{N \times 3})]}{\EE_{x \sim p_x}\EE_{\R'' \sim \unifR} [\N(y; \R'' x, \sigma^2 \II_{N \times 3})]} 
%      \\
%     &= \R' \Dopt(y)
% \end{align}
% where we used the fact that $\unifR$ is uniform so $\R'' = (\R')^{-1}\R$ is also distributed as $\unifR$.

\subsection{The Optimal Conditional Denoiser is $SO(3)$-Equivariant}
\label{sec:optimal-denoiser-equivariant}

Here, we show the that the optimal denoiser $\Dopt$
is indeed \emph{equivariant} under rotations of $y$.

For an arbitrary rotation $\R'$:
\begin{align}
    \Dopt(\R' y; x_0, \sigma) &= 
    \frac{\EE_{\R \sim \unifR} [\R \circ x_0 \ \N(\R'y; \R \circ x_0, \sigma^2 \II_{N \times 3})]}{\EE_{\R \sim \unifR} [\N(\R'y; \R \circ x_0, \sigma^2 \II_{N \times 3})]} 
     \\
    &= 
    \frac{\EE_{\R \sim \unifR} [\R \circ x_0 \ \N(y; (\R')^{-1}\R \circ x_0, \sigma^2 \II_{N \times 3})]}{\EE_{\R \sim \unifR} [\N(y; (\R')^{-1}\R \circ x_0, \sigma^2 \II_{N \times 3})]} 
     \\
    &= 
    \frac{\EE_{\R \sim \unifR} [\R'((\R')^{-1}\R) x_0 \ \N(y; ((\R')^{-1}\R) x_0, \sigma^2 \II_{N \times 3})]}{\EE_{\R \sim \unifR} [\N(y; ((\R')^{-1}\R) x_0, \sigma^2 \II_{N \times 3})]} 
     \\
    &= \frac{\EE_{\R'' \sim \unifR} [\R' \R'' x_0 \ \N(y; \R'' x_0, \sigma^2 \II_{N \times 3})]}{\EE_{\R'' \sim \unifR} [\N(y; \R'' x_0, \sigma^2 \II_{N \times 3})]} 
     \\
    &= \R' \frac{\EE_{\R'' \sim \unifR} [ \R'' x_0 \ \N(y; \R'' x_0, \sigma^2 \II_{N \times 3})]}{\EE_{\R'' \sim \unifR} [\N(y; \R'' x_0, \sigma^2 \II_{N \times 3})]} 
     \\
    &= \R' \Dopt(y; x_0, \sigma)
\end{align}
where we used the fact that $\unifR$ is uniform so $\R'' \equiv (\R')^{-1}\R$ is also distributed as $\unifR$, by the invariance of the Haar measure.

Next, we show that the optimal conditional denoiser $\Dopt$
is indeed \emph{invariant} under rotations of conditioning $x_0$.
For an arbitrary rotation $\R'$:
\begin{align}
    \Dopt(y; \R'x_0, \sigma) &= 
    \frac{\EE_{\R \sim \unifR} [\R \circ \R' x_0 \ \N(y; \R \circ \R' x_0, \sigma^2 \II_{N \times 3})]}{\EE_{\R \sim \unifR} [\N(y; \R \circ \R' x_0, \sigma^2 \II_{N \times 3})]} 
     \\
      &= 
    \frac{\EE_{\R \sim \unifR} [(\R \R') x_0 \ \N(y; (\R  \R') \circ x_0, \sigma^2 \II_{N \times 3})]}{\EE_{\R \sim \unifR} [\N(y; (\R\R') \circ x_0, \sigma^2 \II_{N \times 3})]} 
     \\
      &= 
    \frac{\EE_{\R'' \sim \unifR} [\R'' x_0 \ \N(y; \R'' \circ x_0, \sigma^2 \II_{N \times 3})]}{\EE_{\R \sim \unifR} [\N(y; \R'' \circ x_0, \sigma^2 \II_{N \times 3})]} \\
    &= \Dopt(y; x_0, \sigma)
\end{align}
where we again used the fact that $\unifR$ is uniform so $\R'' \equiv \R  \R'$ is also distributed as $\unifR$, by the invariance of the Haar measure.

\subsection{Rotational Alignment Commutes with Rotational Augmentation}
\label{sec:alignment-with-augmentation}

Here, we show that alignment commutes with the rotation $\Raug$ used for augmentation. In particular, the alignment procedure returns $\Ropt(\Raug \circ y, \Raug \circ x) = \Raug \Ropt(y, x) \Raug^T$.
This is because:
\begin{align}
    \Ropt(\Raug \circ y, \Raug \circ x) &= \argmin_{\R \in SO(3)} \norm{\Raug \circ y - \R \Raug \circ x}  \\
    &= \argmin_{\R \in SO(3)} \norm{y - \Raug^T \R \Raug \circ x}  \\
    \implies \Raug^T \Ropt(\Raug \circ y, \Raug \circ x)\Raug &= \Ropt(y, x)  \\
    \implies \Ropt(\Raug \circ y, \Raug \circ x) &= \Raug \Ropt(y, x) \Raug^T
\end{align}
Thus, on aligning $\Raug \circ x$ to $\Raug \circ y$, we get $\Ropt(\Raug y, \Raug x) \circ (\Raug x) = \Raug \Ropt(y, x) \Raug^T \Raug \circ x = 
\Raug \Ropt(y, x) \circ x$.

\subsection{Averaging an Estimator Induces an Equivalent Matching Loss}
\label{sec:averaging-an-estimator}
Here, we show that averaging an estimator $\Dest$ gives us an equivalent matching loss from the perspective of minimization with respect to $D$.
Formally, for any estimator $\Dest$ we have:
\begin{align}
    \lest(D; \EE_{\R \sim \pRcyx}[\Dest]) = \lest(D; \Dest) + C
\end{align}
where $C$ is a constant that does not depend on $D$.
We have:
\begin{align}
    &\lest(D; \Dest)  \nonumber \\
    &= \EE_{\sigma \sim p_\sigma}\EE_{y\sim p(y|\sigma)}\EE_{x \sim \pxcy}\EE_{\R \sim \pRcyx}[\norm{D(y; \sigma) - \Dest(y; x, \R, \sigma)}^2] \\
    &= \EE_{\sigma}\EE_{y}\EE_{x}\EE_{\R}[\norm{D(y; \sigma) - \Dest(y; x, \R, \sigma)}^2] \\
    &= \EE_{\sigma}\EE_{y}\EE_{x}\EE_{\R}[\norm{D(y; \sigma) - \EE_{\R}[\Dest(y; x, \R, \sigma)] + \EE_{\R}[\Dest(y; x, \R, \sigma)] - \Dest(y; x, \R, \sigma)}^2] \\
    &= \EE_{\sigma}\EE_{y}\EE_{x}\EE_{\R}[\norm{D(y; \sigma) - \EE_{\R}[\Dest(y; x, \R, \sigma)]}^2 + \norm{\EE_{\R}[\Dest(y; x, \R, \sigma)] - \Dest(y; x, \R, \sigma)}^2 \nonumber \\ &+ 2(D(y; \sigma) - \EE_{\R}[\Dest(y; x, \R, \sigma)])^\top(\EE_{\R}[\Dest(y; x, \R, \sigma)] - \Dest(y; x, \R, \sigma))]
\end{align}
where we omit the explicit distributions for clarity.
Now, focusing on the last term:
\begin{align}
    &\EE_\R[2(D(y; \sigma) - \EE_{\R}[\Dest(y; x, \R, \sigma)])^\top(\EE_{\R}[\Dest(y; x, \R, \sigma)] - \Dest(y; x, \R, \sigma))] \\
    &= 2(D(y; \sigma) - \EE_{\R}[\Dest(y; x, \R, \sigma)])^\top \EE_\R[(\EE_{\R}[\Dest(y; x, \R, \sigma)] - \Dest(y; x, \R, \sigma))] \\
    &= 2(D(y; \sigma) - \EE_{\R}[\Dest(y; x, \R, \sigma)])^\top (\EE_{\R}[\Dest(y; x, \R, \sigma)] - \EE_\R[\Dest(y; x, \R, \sigma)]) \\
    &= 2(D(y; \sigma) - \EE_{\R}[\Dest(y; x, \R, \sigma)])^\top \mathbf{0} \\
    &= \mathbf{0}.
\end{align}
as the first term in the product is a constant with respect to $\R$.
Thus,
\begin{align}
    &\lest(D; \Dest) \\
    &= \EE_{\sigma}\EE_{y}\EE_{x}\EE_{\R}[\norm{D(y; \sigma) - \EE_{\R}[\Dest(y; x, \R, \sigma)]}^2 + \norm{\EE_{\R}[\Dest(y; x, \R, \sigma)] - \Dest(y; x, \R, \sigma)}^2] \\
    &= \lest(D; \EE_\R[\Dest]) + \underbrace{\EE_{\sigma}\EE_{y}\EE_{x}\EE_{\R}[\norm{\EE_{\R}[\Dest(y; x, \R, \sigma)] - \Dest(y; x, \R, \sigma)}^2]}_{\text{independent of} \ D}.
\end{align}
as claimed.

\section{For a General Matrix Fisher Distribution}
\label{sec:laplace_details}
We note that the partition function $Z(F)=\int_{SO(3)}\exp(\Tr[F^\top\R])\mathrm{d}\R$ gives all the information we need. In particular, taking derivatives of $Z(F)$ allows us to calculate any necessary moments. For example, 
\begin{align}
    \EE_{\R \sim \MF(\R; F)}[\R] = \frac{\int_{SO(3)}\R \exp(\Tr[F^\top\R])\mathrm{d}\R}{Z(F)} = \frac{\frac{d}{dF} Z(F)}{Z(F)} = \frac{d}{dF} \ln Z(F)
\end{align}
using the trace derivative identity: $\frac{d}{dF} \Tr[F^\top \R] = \R$.

Now, we remove the explicit $\sigma$-dependence of $F = \frac{y^\top x}{\sigma^2}$, by defining $F' = y^\top x$ and $\lambda = \frac{1}{\sigma^2}$, so that $F = \lambda F'$. We are interested in computing $\EE_{\R \sim \MF(\R; \lambda F')}[\R]$ in the limit of $\lambda \to \infty$.

Recall that we can always factorize $F'=USV^\top$ where $U,V\in SO(3)$ and $S=\diag[s_1,s_2,s_3]$ where $s_1\geq s_2\geq |s_3|$, by the Singular Value Decomposition. In particular, we see that: 
\begin{align}
\Tr[F'^\top\R]=\Tr[VS^\top U^\top\R]=\Tr[S^\top U^\top\R V]=\Tr[S^\top(U^\top\R V)]
\end{align}
by the cyclic property of the trace. Since $U,V\in SO(3)$ it follows that $Z(F')=Z(S)$ (by change of variables $\R \to U^\top\R V$) so we can restrict our calculations to the diagonal case .

Laplace's method provides a powerful tool to expand integrals of sharply peaked functions. The key idea is that only the neighborhood of a sharp peak has significant contributions and that such a region can be approximated with a Gaussian distribution. In our case, we seek to apply this method to $Z(\lambda S)$ in the limit of $\lambda\to\infty$.

As $S$ is diagonal, it is easy to see that $\argmax_{\R\sim SO(3)}[\lambda\Tr[S\R]]=\II_{3\times3}$. Hence, a natural parameterization to use is the exponential map expansion of $SO(3)$ which expands around the identity. In this chart, we have parameters $\btheta=(\theta_x,\theta_y,\theta_z)$ and our rotation is given by
\begin{align}\R(\theta_x,\theta_y,\theta_z)=\exp(\theta_x R_x + \theta_y R_y + \theta_z R_z)\end{align}
where $R_x,R_y,R_z$ are the generators of $x,y,z$ rotations. In this parameterization, the Haar measure can be found to be
\begin{align}\mu(\btheta)\mathrm{d}\theta_x\mathrm{d}\theta_y\mathrm{d}\theta_z=\frac{1-\cos(\|\btheta\|)}{4\pi^2\|\btheta\|^2}\mathrm{d}\theta_x\mathrm{d}\theta_y\mathrm{d}\theta_z.\end{align}

Next, we would like to expand the argument $\lambda A(\btheta, S)=\lambda\Tr[S\R(\btheta)]$ around $\btheta=0$. Because this is maximized at $\R=\II$ so $\btheta=0$, we obtain an expression of the form:
\begin{align}
A(\btheta, S)=A_0(S)+ \sum_{ij}A_{2,ij}(S)\theta_i\theta_j+ \sum_{ijk}A_{3,ijk}(S)\theta_i\theta_j\theta_k+\ldots
\end{align}
where the indices $i, j, k$ run over $\{x, y, z\}$.

Define $B(\btheta,S,\lambda) \equiv \exp(\lambda(A(\btheta, S)-A_0(S)- \sum_{ij}A_{2,ij}(S)\theta_i\theta_j))$. We can then write:
\begin{align*}
    \exp(\lambda A(\btheta, S))= & \exp(\lambda A_0(S)+\sum_{ij}\lambda A_{2,ij}(S)\theta_i\theta_j)\exp(\lambda(A(\btheta, S)-A_0(S)- \sum_{ij}A_{2,ij}(S)\theta_i\theta_j))\\
    = & \exp(\lambda A_0(S)+\sum_{ij}\lambda A_{2,ij}(S)\theta_i\theta_j)B(\btheta,S,\lambda).
\end{align*}
To integrate over all of $SO(3)$, we simply need to integrate over the domain $\{|\btheta|<\pi\}$. Hence, we would like to evaluate:
\begin{align}
    \int_{|\btheta|<\pi}e^{\lambda A_0(S)+\sum_{ij}\lambda A_{2,ij}(S)\theta_i\theta_j}B(\btheta,S,\lambda)\mu(\btheta)\mathrm{d}\btheta. \label{eqn:laplace_sep_integral}
\end{align}
Since $\btheta=0$ is a local maxima, $A_{2,ij}(S)$ must be negative definite so the exponential component can be interpreted as a Gaussian. As $\lambda\to\infty$, this Gaussian becomes increasingly peaked; therefore, only the neighborhood around $\btheta=0$ matters. Hence, we can expand $B(\btheta,S,\lambda)\mu(\btheta)$ around $0$ to get:
\begin{align}B(\btheta,S,\lambda)\mu(\btheta)=M_0+\sum_{ij}M_{2,ij}\theta_i\theta_j+\sum_{ijk}M_{3,ijk}(S,\lambda)\theta_i\theta_j\theta_k+\ldots\end{align}
where the contributions up to second order can only come from the expansion of the measure (which has no $S$ or $\lambda$ dependence), and there is no first order term since the measure is symmetric around $0$. Hence, \autoref{eqn:laplace_sep_integral} becomes:
\begin{align}
    \int_{|\btheta|<\pi}e^{\lambda A_0(S)+\sum_{ij}\lambda A_{2,ij}(S)\theta_i\theta_j}\left(M_0+\sum_{ij}M_{2,ij}\theta_i\theta_j+\sum_{ijk}M_{3,ijk}(S,\lambda)\theta_i\theta_j\theta_k+\ldots\right)\mathrm{d}\btheta. \label{eqn:laplace_expanded_integral}
\end{align}

Finally, we note that as $\lambda\to\infty$, the Gaussian part has an increasingly sharp peak. Thus, for the expansion terms in \autoref{eqn:laplace_expanded_integral}, the boundaries of integration matter increasingly less. Hence, replacing the domain $\{|\btheta|<\pi\}$ with the larger domain $\{\btheta \in \RR^3\}$ gives us a good approximation for each of the expanded terms, but also gives us Gaussian integrals which are analytically evaluable.

Finally, we obtain an expression of the form:
\begin{align}Z(\lambda S)=N(S,\lambda)\left(1+L_1(S)\frac{1}{\lambda}+L_2(S)\frac{1}{\lambda^2}+L_3(S)\frac{1}{\lambda^3}+\ldots\right)\end{align}
where $N(S,\lambda)$ is a normalization term. The corresponding expected rotation can be computed as:
\begin{align}
    \EE_{\R\sim\MF(\R;\lambda S)}[\R]  &= \frac{1}{\lambda}\diag\left[\frac{\partial \ln Z(\lambda S)}{\partial s_1},\frac{\partial \ln Z(\lambda S)}{\partial s_1},\frac{\partial \ln Z(\lambda S)}{\partial s_3}\right] \nonumber\\
&= \II+C_1(S)\frac{1}{\lambda}+C_2(S)\frac{1}{\lambda^2}+\ldots
    \label{eqn:expected-R-expand-appendix}
\end{align}
For an arbitrary $F'=USV^\top$, we would then have:
\begin{align}
    \EE_{\R\sim\MF(\R;\lambda F')}[\R]=U\EE_{\R\sim\MF(\R; \lambda S)}[\R]V^\top.
\end{align}
\section{Additional Results in Practice}
\label{sec:additional-results}

Here, we report the RMSD after alignment for the same training runs in \autoref{fig:error-plot-mlp-rmsd-all-frames} and \autoref{fig:error-plot-mlp-rmsd-single-frame}. Note that our estimators are not optimal for this metric; indeed, they minimize the deviation to the optimal denoiser, not to the aligned ground truth $x$. The complication is that computing the optimal denoiser is not practical in a training setup.

\begin{figure}[h]
    \centering
    \includegraphics[width=0.95\textwidth]
    {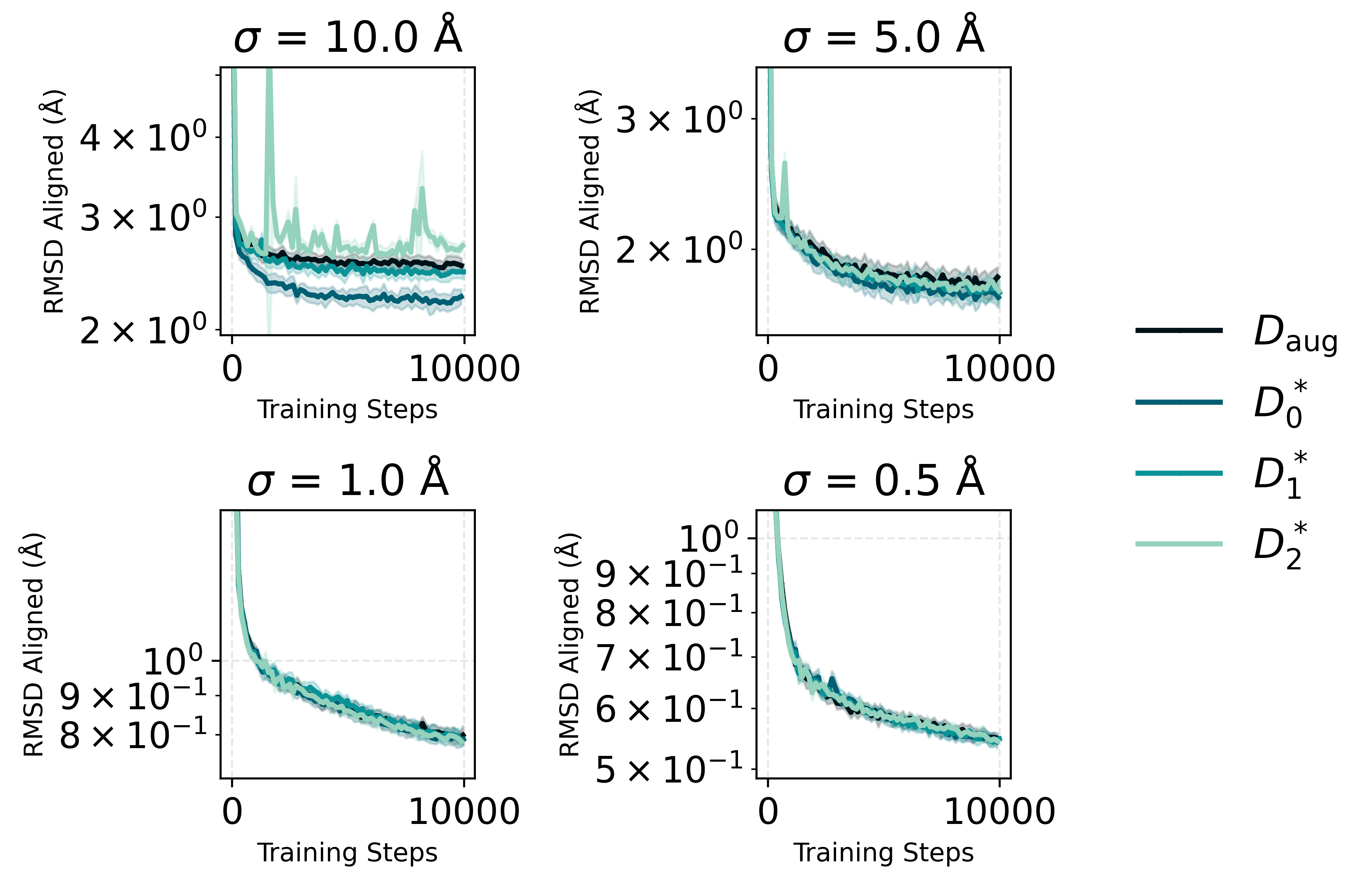}
    \caption{Training progress for the MLP, as measured by RMSD to ground-truth $x$ after alignment, when trained using $\Daug$, $\Dopt_0$, $\Dopt_1$ and $\Dopt_2$. $x$ is sampled from all $50000$ frames of a molecular dynamics simulation for the \texttt{AEQN} peptide.}
    \label{fig:error-plot-mlp-rmsd-aligned-all-frames}
\end{figure}

\begin{figure}[h]
    \centering
    \includegraphics[width=0.95\textwidth]
    {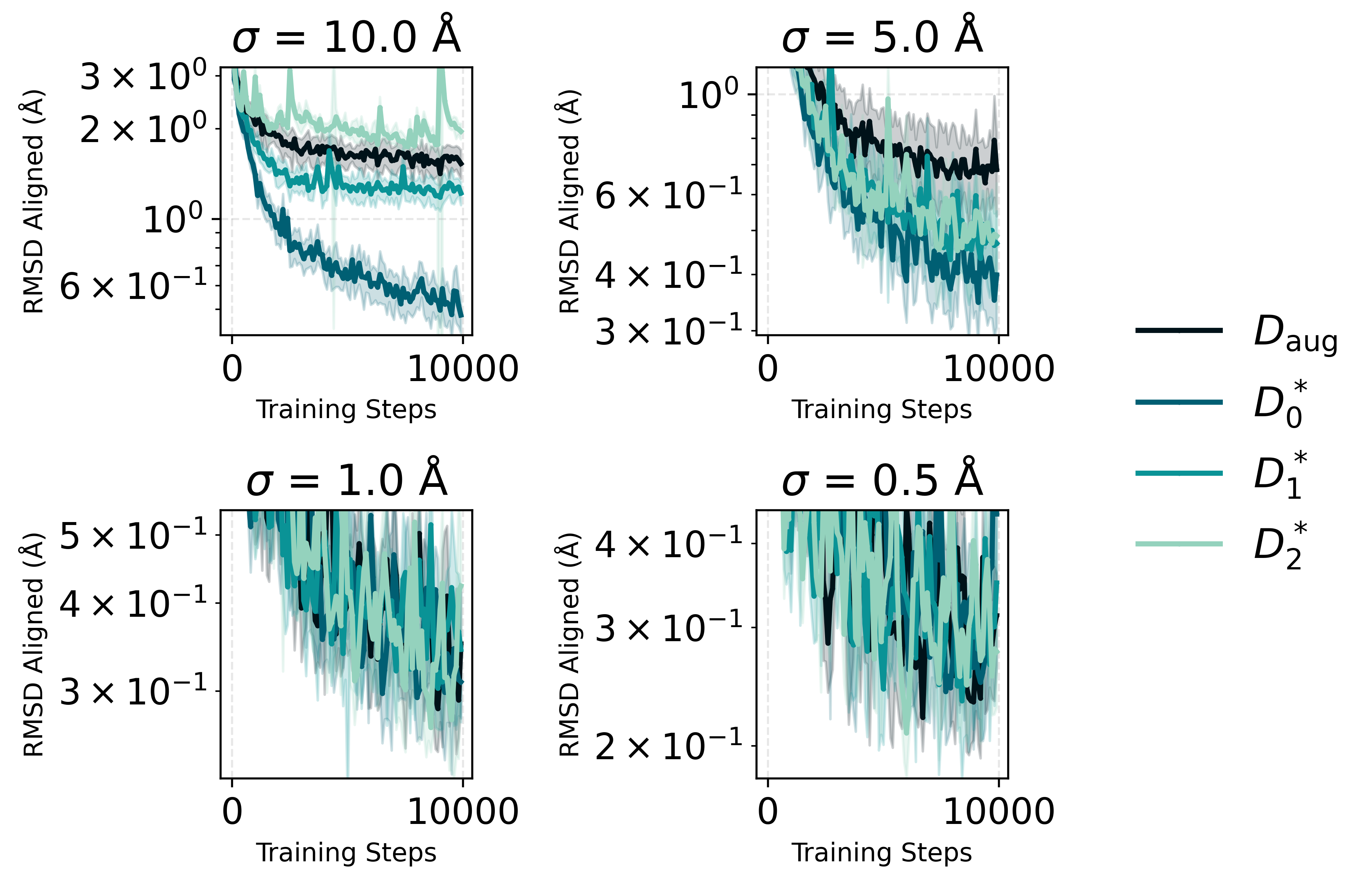}
    \caption{Training progress for the MLP, as measured by RMSD to ground-truth $x$ after alignment, when trained using $\Daug$, $\Dopt_0$, $\Dopt_1$ and $\Dopt_2$. $x$ is fixed as the first frame in the molecular dynamics simulation for the \texttt{AEQN} peptide.}
    \label{fig:error-plot-mlp-rmsd-aligned-single-frame}
\end{figure}

\end{document}